\documentclass[review]{elsarticle}
\usepackage{packages}
\begin{document}
\begin{frontmatter}
	% !TEX root = ../main.tex
\title{The Multiscale Bowler-Hat Transform\\ for Blood Vessel Enhancement in Retinal Images}
\author[rvt]{\c{C}i\u{g}dem~Sazak}
\ead{cigdem.sazak@durham.ac.uk}
\author[els]{Carl~J.~Nelson}
\ead{chas.nelson@glasgow.ac.uk}
\author[rvt]{Boguslaw~Obara\corref{cor1}}
\ead{boguslaw.obara@durham.ac.uk}
\cortext[cor1]{Corresponding author}
\address[rvt]{Department of Computer Science Durham University, UK}
\address[els]{School of Physics and Astronomy Glasgow University, UK}
	% !TEX root = ../main-elsevier.tex
% % % % % % % % % % %
% ABSTRACT
% % % % % % % % % % %
\begin{abstract}
	Enhancement, followed by segmentation, quantification and modelling, of blood vessels in retinal images plays an essential role in computer-aided retinopathy diagnosis. In this paper, we introduce the bowler-hat transform method, a new approach based on mathematical morphology for vessel enhancement. The proposed method combines different structuring elements to detect innate features of vessel-like structures. We evaluate the proposed method qualitatively and quantitatively and compare it with the existing, state-of-the-art methods using both synthetic and real datasets. Our results establish that the proposed method achieves high-quality vessel-like structure enhancement in both synthetic examples and clinically relevant retinal images, and is shown to be able to detect fine vessels while remaining robust at junctions.
\end{abstract}

\end{frontmatter}
%------------------------------------------------------------------------------%
% !TEX root = ../main-elsevier.tex
\section{Introduction}\label{sec:intro}

Many biomedical images contain vessel-like structures, such as blood vessels or cytoskeletal networks~\cite{Bibiloni2016}. Automated extraction of these structures and their connected network is often an essential step in quantitative image analysis and computer-aided diagnostic pipelines. For example, automated retinal vessel extraction is used for diagnosis, screening, and evaluation in a wide range of retinal diseases, including diabetes and arteriosclerosis~\cite{Fraz2012}.

However, for a multitude of reasons, \eg{} noisy image capture, sample/patient variability, low contrast scenarios, etc., biomedical imaging modalities may suffer from poor quality.
As such, standard image segmentation methods are not able to robustly detect vessel-like structures, and therefore some form of vessel-like structure enhancement is required~\cite{Bibiloni2016}.

A wide range of vessel enhancement methods have been proposed (see~\cite{Fraz2012} and~\cite{Bibiloni2016} for a recent review). These include Hessian~\cite{frangi1998multiscale,Su2012,Su2014}, Phase Congruency Tensor~\cite{obara2012contrast,obara2012coherence}, mathematical morphology~\cite{ZK2001,Su2014,lu2016vessel}, adaptive histogram equalisation~\cite{Pisano1998} based approaches and many others~\cite{Feng2007,FBRetal2012,bankhead2012fast,Azzopardi2013,nguyen2013effective,Sigurosson2014,CCSetal2016,Merveille2017}.

However, many of these methods still have considerable issues when faced with variations in contrast, high levels of noise, variation in image features (\eg{} lines vs junctions; retention of network connectivity), and complexity of method parameter space.

\subsection{Contribution and Organisation}

In this paper, we introduce a new enhancement method for vessel-like structures based on mathematical morphology, which exploits a key shape property of vessel-like structures: elongation. The proposed method, called the bowler-hat transform, has been qualitatively and quantitatively validated and compared with state-of-the-art methods using a range of synthetic data and publicly available retinal image datasets. 
The obtained results show that the proposed method achieves high-quality, vessel-like structure enhancement in both synthetic examples and clinically relevant retinal images. The method is suitable for a range of biomedical image types without needing prior training or tuning. Finally, we have made the implementation of our approach available online, along with source code and all test functions.

The rest of this paper is organised as follows. In~\Cref{sec:related}, we introduce existing vessel-like structure enhancement methods and highlight their known limitations. ~\Cref{sec:methods} introduces and explains the proposed bowler-hat transform,~\Cref{sec:results} presents validation experiments and results on synthetic and real data. Finally, in~\Cref{sec:Conclusion}, we discuss the results and future work.\vspace{-10pt}

% -------------------------------------------------------------------------
\section{Related Work}\label{sec:related}

Since vessel-like structures in images can appear in different sizes, a multi-scale concept for vessel enhancement and segmentation has been extensively investigated in the literature~\cite{frangi1998multiscale,sato19973d,krissian2000model,koller1995multiscale}.
In such a concept, we are embedding the original image into a family of increasingly different scales, where the fine-scale details are successively suppressed. This multi-scale image representation is generally obtained by use of Gaussian filters or their derivatives, with different scales or varying the size of the structuring element correspondingly in each scale in mathematical morphology theory.\vspace{-10pt}

\subsection{Hessian-based Methods}\label{subsec:hessian}

In~\cite{frangi1998multiscale}, Frangi et al. introduce a novel Hessian-based multi-scale concept for 2D curvilinear/3D tubular structure enhancement in images. They construct the Hessian matrix using second-order Gaussian derivatives. The eigenvectors and eigenvalues of the Hessian matrix then define the principal directions of local image features. These can then be combined to form different measures of vesselness or blobness~\cite{Bibiloni2016} in biomedical images.\vspace{-10pt}

\subsubsection{Vesselness}\label{subsec:ves}

The vesselness measure is proportional to the ratio of the eigenvalues~\cite{frangi1998multiscale}. If the magnitude of both eigenvalues is small, \ie{} the local image structure is likely to be background, then the vesselness measure is small. If one eigenvalue is small and the other large then the local structure is likely to be vessel-like, and the vesselness measure is large. Finally, if both eigenvalues are high, then the structure is likely to be blob-like, and the vesselness measure is again small.

This approach, however, leads to a failure at the intersection of vessels as both eigenvalues have similarly large values leading to a vesselness measure close to zero. Thus, vessel-like structures can be lost at junctions and therefore vessels network connectivity may be lost~\cite{krissian2003multiscale}. An extension of this approach can be found in~\cite{Su2014} where a multi-scale morpho-Gaussian filter is combined with multi-scale Hessian measurement to enhance the curvilinear features and reduce noise. 

\subsubsection{Neuriteness}\label{subsec:neurite}

As an alternative to vesselness, Meijering and colleagues~\cite{meijering2004design} introduce the neuriteness measure to enhance low contrast and highly inhomogeneous neurites in bioimages. Using a modified Hessian, with a tuning parameter, and a different combination of eigenvalues, neuriteness infers a putative neurite in every pixel of the image that has a non-zero value. Background intensity discontinuities that are immune to first order derivatives are suppressed by the use of second order derivatives.

A major failing for the neuriteness measure is that background noise signals are enhanced as if they are curvilinear structures. In the original paper~\cite{meijering2004design} this is solved with a tracing stage; however, as an enhancer only, this can cause serious problems for further analysis. The neuriteness measure also leads to a failure at the intersection of vessels as both eigenvalues have similarly large values leading to a neuriteness measure close to zero.
A further example of their work is found in~\cite{smafield2015automatic}.

\subsubsection{Regularized Volume Ratio}\label{subsec:volrat}

Recently, Jerman and colleagues~\cite{JPLetal2016} propose a new Hessian-based vessel enhancement method, which is able to resolve the drawbacks found in most of the previous Hessian-based methods:
\begin{enumerate*}
	\item eigenvalues are non-uniform throughout an elongated or rounded structure that has uniform intensity;
	\item eigenvalues vary with image intensity; and
	\item enhancement is not uniform across scales.
\end{enumerate*}
To address such drawbacks, a modified volume ratio is introduced to ensure method robustness to low magnitude intensity changes in the image. A major issue of this method is the false vessel effect, as also shown in~\Cref{fig:hrf:vr} in~\Cref{subsubsec:real:compare} noise sensitivity. 

\subsection{Phase Congruency Tensor-based Methods}\label{subsec:pc}

A major issue with many image enhancement methods is that they depend on image intensity and, therefore, fine, and usually lower intensity, vessels may be missed. To address this issue a contrast-independent method, based on Phase Congruency (PC), was introduced in~\cite{kovesi2003phase}. This approach builds upon the idea of phase congruency, which looks for image features in the Fourier domain. 

The development of a contrast-independent, image enhancement measurement built upon PC has been shown in~\cite{obara2012contrast}. The Phase Congruency Tensor (PCT) is built upon PC principles, but the tensor is decomposed. The calculated eigenvalues are then used in the same way as Hessian eigenvalues (see~\Cref{subsec:ves,subsec:neurite}), to define PCT vesselness and PCT neuriteness measures. An extension of this method into 3D has recently been shown in~\cite{Sazak2017}.

A major drawback of the PC-based concept is the complexity of its parameter space. Moreover, as with Hessian-based measures, the PCT-based measures also lead to a failure at the intersection of vessels as both eigenvalues have similar, large values leading to PCT-based vesselness and neuriteness measures close to zero. \vspace{-10pt}

\subsection{Adaptive Histogram Equalisation-based Methods}\label{subsec:clahe}

Contrast Limited Adaptive Histogram Equalisation (CLAHE)~\cite{Pisano1998}, originally developed for speculations enhancement in mammograms, is widely used for vessel enhancement. In this simple, histogram-based method an image is first divided into small regions, each of which then undergoes a histogram equalisation. To avoid over-enhancement of noise, a contrast limiting procedure is applied between regions. Further development of this method is demonstrated in~\cite{Zhao2014} where CLAHE is combined with an anisotropic diffusion filter to smooth the image and preserve vessel boundaries. A major drawback of this method is the noise sensitivity. 

\subsection{Wavelet Transform-based Enhancement Methods}\label{subsuce:wavelet}

Bankhead and colleagues~\cite{bankhead2012fast} propose the use of wavelets for vessel enhancement and segmentation. They calculate an isotropic, undecimated wavelet transform using the cubic B-spline mother wavelet, and employ the coefficients to the threshold steps for enhancement, followed by vessel segmentation. Further improvement of this approach is demonstrated in~\cite{Zhang2017} where multi-orientation and multi-scale features from the vessel filtering and the wavelet transform stages are combined and then used for training the random forest classifier.
A major drawback of this method is the complexity of its parameter space. 

\subsection{Line Detector-based Enhancement Methods}\label{subsec:linedetector}

Vessel-like feature enhancement has also been done using multi-scale line detectors~\cite{nguyen2013effective}. The basic line response, identified by subtraction of average value and the maximum value of each pixel, is computed at 12 different line directions. A major drawback of this method is at crossover points, where the method produces `false vessels' by merging nearby vessels. 
Further improvement of this method is demonstrated in~\cite{hou2014automatic} where a linear combination of all the line responses at varying scales is proposed to produce the final enhancement and segmentation.

\subsection{Mathematical Morphology-based Enhancement Methods}\label{subsec:morph}

Zana and Klein~\cite{ZK2001} proposed a novel method which combines morphological transforms and cross-curvature evaluation for vessel-like structure enhancement and segmentation. This method relies on the assumption that vessels are linear, connected and have smooth variations of curvature along the peak of the feature. First, a sum of top hats is calculated using linear structuring elements with the single size (15-pixels long) at different orientations, and after enhancement step, a curvature measure is calculated using a Laplacian of Gaussian, and finally, both of them are combined to reduce noise and enhance vessel-like structures in an image. 
Further improvement of this method is demonstrated in~\cite{Su2014,Sigurosson2014,lu2016vessel}. 
In particular, in~\cite{Sigurosson2014}, an advanced morphological directional filter called path openings is linked with data fusion based on fuzzy set theory. This approach has four steps; First is preprocessing, where the image undergoes histogram equalisation, and then Gaussian filtering to improve the effectiveness of the second step. The second step involves feature extraction by detection of local minima and edges in the image. The third step preserves connected vessels and suppresses noise by path opening, and the final step combines the features and possible paths into a fuzzy classification problem - identifying pixels as likely vessels or likely background.

And most recently, in~\cite{lu2016vessel}, a multi-scale morphological top hats transform is combined with Gabor and a matched filter. A major issue with this method is that it is quite slow and sensitive to noise. 

\subsection{Limitations and Challenges}\label{subsec:limits}

Many existing vessel-like structure enhancement methods still have substantial issues when faced with variations in contrast (low-accuracy enhancement), high levels of noise (introduction of 'false vessels' effect), dealing with junctions/bends (suppression of disk-like structures; vessels network connectivity is lost), large image size (high computing time), and complexity of parameter space.\vspace{-10pt}

%------------------------------------------------------------------------------%
% !TEX root = ../main-elsevier.tex
% FIGURE
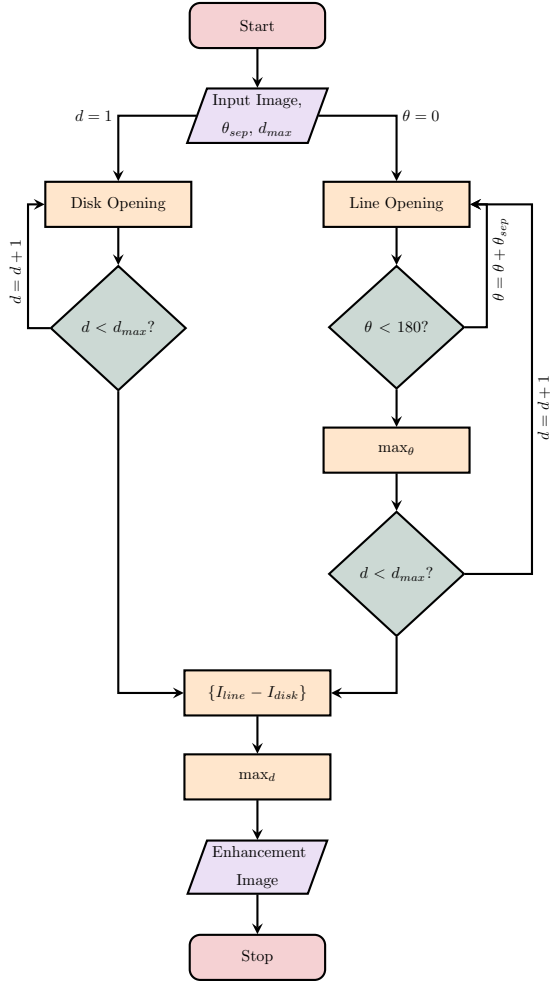
\begin{figure}[th!]
	\tikzstyle{startstop} = [rectangle, rounded corners, minimum width=3cm, minimum height=1cm,text centered, draw=black, fill=red!20]
	\tikzstyle{io} = [trapezium, trapezium left angle=70, trapezium right angle=110, minimum width=3cm, minimum height=1cm, text centered, text width=2cm, draw=black, fill=blue!20]
	\tikzstyle{process} = [rectangle, minimum width=3cm, minimum height=1cm, text centered,text width=3cm, draw=black, fill=orange!20]
	\tikzstyle{decision} = [diamond, minimum width=3cm, minimum height=1cm, text centered,text width=2cm, draw=black, fill=green!20]
	\tikzstyle{arrow} = [thick,->,>=stealth]
	\centering
	\begin{tikzpicture}[node distance=2cm,thick,scale=0.6, every node/.style={scale=0.6}]
		% Start
		\node (start) [startstop] {Start};
		\node (input) [io, below of=start] {Input Image,  $\theta_{sep}$, $d_{max}$};
		% Disk route
		\node (proDisk) [process, below left=0.5cm and 0.5cm of input] {Disk Opening};
		\node (decDiskR) [decision, below=0.5cm of proDisk] {$d<d_{max}$?};
		% Line route
		\node (proLine) [process, below right=0.5cm and 0.5cm of input] {Line Opening};
		\node (decLineT) [decision, below=0.5cm of proLine] {$\theta<180$?};
		\node (proMaxT) [process, below=0.5cm of decLineT] {$\max_\theta$};
		\node (decLineR) [decision, below=0.5cm of proMaxT] {$d<d_{max}$?};
		% Diff & Finish
		\node (proDiff) [process, below=7cm of input] {$\{I_{line} - I_{disk}\}$};
		\node (proMaxD) [process, below=0.5cm of proDiff] {$\max_d$};
		\node (out1) [io, below of=proMaxD] {Enhancement Image};
		\node (stop) [startstop, below of=out1] {Stop};
		% Arrows
		\draw [arrow] (start) -- (input);
		\draw [arrow] (input) -| node[anchor=east] {$d=1$}(proDisk);
		\draw [arrow] (proDisk) -- (decDiskR);
		\draw [arrow] (decDiskR) |- (proDiff);
		\draw [arrow] (decDiskR) --++  (-2,0)  |- node [rotate=90,anchor = south, xshift = -1.5cm]{$d=d+1$}(proDisk);
		\draw [arrow] (input) -| node[anchor=west] {$\theta=0$}(proLine);
		\draw [arrow] (proLine) -- (decLineT);
		\draw [arrow] (decLineT) -- (proMaxT);
		\draw [arrow] (decLineT) --++  (2,0) node [rotate=90, anchor = north,xshift = 1.5cm]{$\theta=\theta+\theta_{sep}$} |- (proLine);
		\draw [arrow] (proMaxT) -- (decLineR);
		\draw [arrow] (decLineR) |- (proDiff);
		\draw [arrow] (decLineR) --++  (3,0) |- node [rotate = 90, anchor = north,xshift = -4.5cm]{$d=d+1$}(proLine);
		\draw [arrow] (proDiff) -- (proMaxD);
		\draw [arrow] (proMaxD) -- (out1);
		\draw [arrow] (out1) -- (stop);
	\end{tikzpicture}
	\caption{Flow chart of the bowler-hat.}\label{fig:flow}
\end{figure}

\section{Method}\label{sec:methods}

In this section, we introduce our novel, mathematical morphology-based method for vessel-like structure enhancement in images: the bowler-hat transform. We highlight the key concepts that allow this method to address the major drawbacks of existing, state-of-the-art, methods.

\subsection{Mathematical Morphology}

Morphological operations are a set of non-linear filtering methods formed through a combination of two basic operators: dilation and erosion.

Dilation, ($\oplus$), for a given pixel in a greyscale image, $I(\mathbf{p})$, can be described as the maximum of the points in the weighted neighbourhood described by the structuring element $b(\mathbf{p})$, and mathematically:
\begin{equation}
(I \oplus b)(\mathbf{p}) = \sup_{\mathbf{x} \in E}[I(\mathbf{x})+b(\mathbf{p}-\mathbf{x})],
\end{equation}
where `$\sup$' is the supremum and $\mathbf{x} \in E$ denotes all points in Euclidean space within the image. Likewise, we mathematically describe the erosion ($\ominus$), as the minimum of the points in the neighbourhood described by the structuring element:
\begin{equation}
(I \ominus b)(\mathbf{p}) = \inf_{\mathbf{x} \in E}[I(\mathbf{x})+b(\mathbf{p}-\mathbf{x})],
\end{equation}
where `$\inf$' is the infimum. Dilation is able to expand bright areas and reduce dark areas, whilst erosion expands dark areas reducing bright areas as detailed in~\cite{haralick1987image}.
From these two operators we can define two commonly used morphological filters:
\begin{align}
	\text{opening}:    & \quad (I \circ b)(\mathbf{p}) = ((I \ominus b) \oplus b)(\mathbf{p})   \\
	\text{closing}:    & \quad (I \bullet b)(\mathbf{p}) = ((I \oplus b) \ominus b)(\mathbf{p})
\end{align}
where an opening ($\circ$) will preserve dark features and patterns, suppressing bright features, and a closing ($\bullet$) will preserve bright features whilst suppressing dark patterns.

\subsection{Proposed Method}\label{subsec:proposed}

\Cref{fig:flow} presents a flow diagram of the proposed method which combines the outputs of morphological operations upon an image carried out with two different banks of structural elements: one bank of disk elements with varying radii, and one bank of line elements with varying radii and rotation.
The bowler-hat transform is named after the bank of disk elements (forming the bowl) and the bank of line elements (forming the brim).
For a given greyscale input image, $I$, we carry out a series of morphological openings with a bank of disk-shaped structuring elements, $b_d$ of diameter $d\in{[1,d_{max}]}$ pixels, where $d_{max}$ is the expected maximum vessel size and user-defined parameter. This produces a stack of images, for all $d$, such that
\begin{equation}
\{I_{disk}\} = \{I \circ b_d \colon \forall d \in [1,d_{max}]\}.
\end{equation}
In each $I_{disk}$ image, vessel segments wider than $d$ remain and those segments smaller than $d$ are removed.

We also produce a similar stack of images using a bank of line-shaped structuring elements, $b_{d,\theta}$; each line-shaped is of length $d\in{[1,d_{max}}]$, with a width of $1$ pixel, and orientation $\theta\in{[0,180)}$, $\theta_{sep}$ is angle step.

As a result, vessel segments that are longer than $d$ and along the direction defined by $\theta$ will remain, and those shorter than $d$ or along the direction defined by $\theta$ will be removed. For each line length $d$ we produce a stack of images for all orientations defined by $\theta\in{[0,180)}$. Then, for each $d$, we calculate a single image, $I_{line}$ as a pixel-wise maximum of the stack such that
\begin{equation}
\{I_{line}\} = \{\max_\theta(\{I \circ b_{d,\theta} \colon \forall \theta\}) \colon \forall d \in [1,d_{max}]\}.
\end{equation}

These two stacks, $\{I_{disk}\}$ and $\{I_{line}\}$, are then combined by taking the stack-wise difference, the difference between the maximum opening with a line of length $d$ across all angles and an image formed of opening with a disk of size $d$, to form the enhanced image. The final enhanced image is then formed from maximum difference at each pixel across all stacks,
\begin{equation}
I_{enhanced} = \max_d(\lvert I_{line} - I_{disk}\rvert).
\end{equation}

Pixels in the background, \ie{} dark regions, will have a low value due to the use of openings; pixels in the foreground of blob-like structures will have a low value as the differences will be minimal, \ie{} similar values for disk-based and line-based openings; and pixels in the foreground of vessel-like structures will have a high value, \ie{} large differences between longer line-based openings and disk-based openings.

The combination of line and disk elements gives the proposed method a key advantage over the existing methods. Given an appropriate $d_{max}$, \ie{} larger than any vessels in the image, a junction should appear bright like those vessels joining that junction, something that many other vessel enhancement methods fail to do. This is due to the ability to fit longer line-based structural elements within the junction area. As a result, the vessels network stays connected when enhanced and segmented, especially at junctions.

In \Cref{sec:results}, we demonstrate, qualitatively and quantitatively, the key advantages of the bowler-hat transform over the existing, state-of-the-art vessel-like structure enhancement methods.

\subsection{Implementation and Computation Time}\label{subsec:imp}

All codes were implemented and written in MATLAB 2016b~\cite{matlab} on Windows 8.1 Pro 64-bit PC running an Intel Core i7-4790 CPU (3.60 GHz) with 16GB RAM. The source code is available in a GitHub repository (\textbf{TBC}).

The average computation time for the proposed method is 3.8 seconds for DRIVE image and 4.9 seconds for STARE image. Please make a note that the proposed method has been implemented and tested in Matlab, however, C++ implementation could be much faster.
%------------------------------------------------------------------------------%
% !TEX root = ../main-elsevier.tex
\section{Results}\label{sec:results}

In this section, the proposed method is qualitatively and quantitatively validated and compared with the existing state-of-the-art methods using synthetic and clinically relevant, retinal image datasets, with human-annotated ground truths, and other biomedical images.

As with any image processing method, an understanding of how the parameters involved affect the result is essential. In general, we have found the bowler-hat transform to be robust, usually requiring 10--12 $\theta$ orientations for line structuring element and the size of the disk/line structuring element $d$ to be greater than the thickest vessel structure in an image.

% FIGURE
\input{./tex/figure-profile}
% FIGURE
\input{./tex/figure-compare}

\subsection{Profile Analysis}

The effect of the vessel enhancement methods on a cross-section profile of the enhanced vessels is demonstrated using a simple vessel-like structure in a synthetic image, see~\Cref{fig:profile}. For a fair comparison, all enhanced images were normalised such that the brightest pixel in the whole image has a value of $1$ and the darkest a value of $0$. As the figure clearly shows, the enhancement methods tend to expand or shrink the vessel-like structures. 
Moreover, while the Hessian-based methods have an enhanced signal at the center of the vessel, \ie{} a peak value of one at the vessels centre-line, their value quickly drops off and decreases the perceived thickness of the vessel. Contrariwise, PCT-based methods do not necessarily peak at the vessel centre, but their response does not drop off quickly, and they maintain a higher response through to the edges of the vessel, \ie{} the perceived thickness is not decreased. The proposed method has both these benefits: a maximal peak value at the vessel centre-line and an enhanced response to the edges of the vessel. As a result, reliable vessel thickness can be captured. \vspace{-10pt}

\subsection{Response to Vessels, Intersections, and Blobs}\label{subsubsec:compare}

~\Cref{fig:compare} presents a qualitative comparison between the proposed method and the state-of-the-art methods when applied to synthetic images and real images with vessel-like, intersection-like, and blob-like structures. Key issues that occur across the state-of-the-art methods include defects at junctions (purple arrows), noise enhancement, tip artefacts (orange arrows) and loss of signal (yellow arrows). These issues are all absent with our proposed method. \vspace{-10pt}

% FIGURE
\renewcommand\sfigure{0.31}
\begin{figure}[h!]
	\begin{subfigure}{\sfigure\linewidth}
		\begin{tikzpicture}
		\begin{axis}[
		width = 1.2\linewidth,
		xlabel={Peak Signal-to-Noise Ratio (PSNR)},
		ylabel={Area Under ROC (Normalised)},
		ylabel near ticks,
		label style={font=\tiny},
		tick label style={font=\tiny},
		ytickmin=0, ymax=1.1,
		xtickmin=0, xtickmax=25, x dir=reverse,
		enlargelimits=false,
		legend entries = {Raw image, Vesselness, CLAHE,Zana's top-hat, Neuriteness, PCT vesselness, PCT neuriteness, Wavelet, Line detector, Volume ratio, Bowler-hat},
		legend columns = 4,
		legend style={font=\footnotesize,line width=2pt, draw=none,},% cells={anchor=west}},
		legend to name=leg2,
		grid=major, % Display a grid
		grid style={dashed,gray!30}, % Set the style
		]
		\addplot[color=black,loosely dashed] table [x index=0, y index=1, col sep=comma] {image/datFiles/gaussian.dat};%Raw Image
		\addplot[color=brewerDark1] table [x index=0, y index=3, col sep=comma] {image/datFiles/gaussian1.dat};%Vesselness
		\addplot[color=brewerDark6] table [x index=0, y index=8, col sep=comma] {image/datFiles/gaussian.dat};%clahe
		\addplot[color=brewerDark5] table [x index=0, y index=7, col sep=comma] {image/datFiles/gaussian1.dat};%tophat
		\addplot[color=brewerDark2] table [x index=0, y index=4, col sep=comma] {image/datFiles/gaussian.dat};%neuriteness
		\addplot[color=brewerDark3] table [x index=0, y index=5, col sep=comma] {image/datFiles/gaussian.dat};%pct vess
		\addplot[color=brewerDark4] table [x index=0, y index=6, col sep=comma] {image/datFiles/gaussian.dat};%pct neurite
		\addplot[color=brewerDark8] table [x index=0, y index=10, col sep=comma] {image/datFiles/gaussian.dat};%wavelet
		\addplot[color=green!70] table [x index=0, y index=11, col sep=comma] {image/datFiles/gaussian1.dat};%line detector
		\addplot[color=brewerDark7] table [x index=0, y index=9, col sep=comma] {image/datFiles/gaussian.dat};%vr
		\addplot[color=black] table [x index=0, y index=2, col sep=comma] {image/datFiles/gaussian.dat};%proposed
		\end{axis}
		\end{tikzpicture}
		\caption{Additive Gaussian Noise}\label{fig:psnr:gauss}
	\end{subfigure}
	\begin{subfigure}{\sfigure\linewidth}
		\begin{tikzpicture}
		\begin{axis}[
		width =1.2\linewidth,
		xlabel={Peak Signal-to-Noise Ratio (PSNR)},
		ylabel={Area Under ROC (Normalised)},
		ylabel near ticks,
		label style={font=\tiny},
		tick label style={font=\tiny},
		ytickmin=0, ymax=1.1,
		xtickmin=0, xtickmax=25, x dir=reverse,
		enlargelimits=false,
		grid=major, % Display a grid
		grid style={dashed,gray!30}, % Set the style
		]
		\addplot[color=black,loosely dashed] table [x index=0, y index=1, col sep=comma] {image/datFiles/speckle.dat};%raw
		\addplot[color=brewerDark1] table [x index=0, y index=3, col sep=comma] {image/datFiles/speckle.dat};%vesse
		\addplot[color=brewerDark6] table [x index=0, y index=8, col sep=comma] {image/datFiles/speckle.dat};%clahe
		\addplot[color=brewerDark5] table [x index=0, y index=7, col sep=comma] {image/datFiles/speckle.dat};%tophat
		\addplot[color=brewerDark2] table [x index=0, y index=4, col sep=comma] {image/datFiles/speckle.dat};%neurite
		\addplot[color=brewerDark3] table [x index=0, y index=5, col sep=comma] {image/datFiles/speckle.dat};%pct ves
		\addplot[color=brewerDark4] table [x index=0, y index=6, col sep=comma] {image/datFiles/speckle.dat};%pct neurite
		\addplot[color=brewerDark8] table [x index=0, y index=10, col sep=comma] {image/datFiles/speckle.dat};%wavelet
		\addplot[color=green!40] table [x index=0, y index=11, col sep=comma] {image/datFiles/speckle.dat};%line detector
		\addplot[color=brewerDark7] table [x index=0, y index=9, col sep=comma] {image/datFiles/speckle.dat};%vr
		\addplot[color=black] table [x index=0, y index=2, col sep=comma] {image/datFiles/speckle.dat};%proposed
		\end{axis}
		\end{tikzpicture}
		\caption{Speckle Noise}\label{fig:psnr:speck}
	\end{subfigure}
	\begin{subfigure}{\sfigure\linewidth}
		\begin{tikzpicture}
		\begin{axis}[
			width =1.2\linewidth,
		xlabel={Peak Signal-to-Noise Ratio (PSNR)},
		ylabel={Area Under ROC (Normalised)},
		ylabel near ticks,
		label style={font=\tiny},
		tick label style={font=\tiny},
		ytickmin=0, ymax=1.1,
		xtickmin=0, xtickmax=25, x dir=reverse,
		enlargelimits=false,
		grid=major, % Display a grid
		grid style={dashed,gray!30}, % Set the style
		]
		\addplot[color=black,loosely dashed] table [x index=0, y index=1, col sep=comma] {image/datFiles/pepper.dat};%raw
		\addplot[color=brewerDark1] table [x index=0, y index=3, col sep=comma] {image/datFiles/pepper.dat};%vess
		\addplot[color=brewerDark6] table [x index=0, y index=8, col sep=comma] {image/datFiles/pepper.dat};%clahe
		\addplot[color=brewerDark5] table [x index=0, y index=7, col sep=comma] {image/datFiles/pepper.dat};%tophat
		\addplot[color=brewerDark2] table [x index=0, y index=4, col sep=comma] {image/datFiles/pepper.dat};%neuriteness
		\addplot[color=brewerDark3] table [x index=0, y index=5, col sep=comma] {image/datFiles/pepper.dat};%pct vess
		\addplot[color=brewerDark4] table [x index=0, y index=6, col sep=comma] {image/datFiles/pepper.dat};%pct neurite
		\addplot[color=brewerDark8] table [x index=0, y index=10, col sep=comma] {image/datFiles/pepper.dat};%Wavelet
		\addplot[color=green!70] table [x index=0, y index=11, col sep=comma] {image/datFiles/pepper.dat};%line detector
		\addplot[color=brewerDark7] table [x index=0, y index=9, col sep=comma] {image/datFiles/pepper.dat};%vr
		\addplot[color=black] table [x index=0, y index=2, col sep=comma] {image/datFiles/pepper.dat};%proposed
		\end{axis}
		\end{tikzpicture}
		\caption{Salt and Pepper Noise}\label{fig:psnr:pepp}
	\end{subfigure}
	\begin{subfigure}{\linewidth}
		\centering
		\pgfplotslegendfromname{leg2}
	\end{subfigure}
	\caption{The bowler-hat transform is robust against additive Gaussian noise but susceptible to speckle and salt\&pepper. Mean AUC for the input image and the image enhanced by bowler-hat and by the state-of-the-art methods with different peak signal-to-noise ratios (PSNRs) for three different noise types: (\subref{fig:psnr:gauss}) additive Gaussian noise, (\subref{fig:psnr:speck}) multiplicative Gaussian noise, and (\subref{fig:psnr:pepp}) salt and pepper noise (see legend for colours).}\label{fig:psnr}
\end{figure}
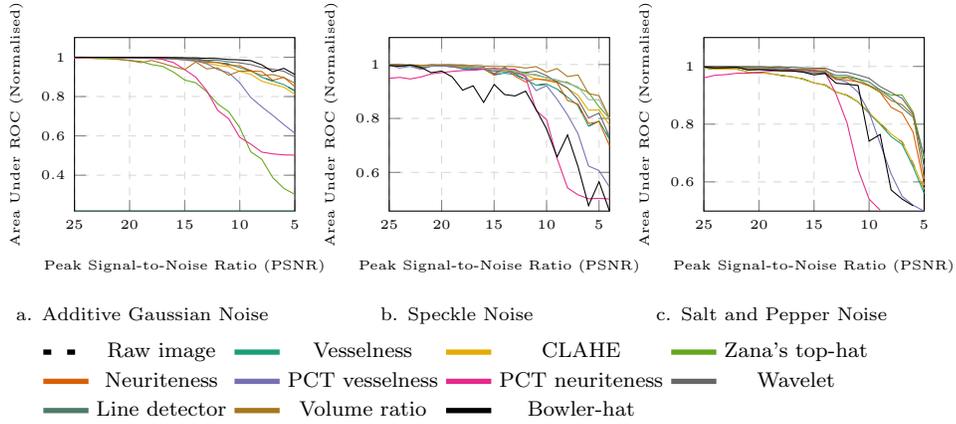

\subsection{Response to Noise}\label{subsubsec:noise}

To test how the state-of-the-art enhancement methods and the proposed method behave with the different level and type of the noise, a noisy synthetic image that includes a single vessel-like structure was used. We generate a noisy image by optimising the noise generation parameters to achieve a target PSNR by using a genetic optimisation algorithm. We then examine the enhancement methods by increasing the noise level and then calculating the AUC values for each level of noise and each comparator method. ~\Cref{fig:psnr} shows the effect of three different noise types on the proposed and state-of-the-art methods. Given that the proposed method has no built-in noise suppression, it is unsurprising that the effect of noise on the enhanced image is in-line with the raw image. We note that the method is weakest in response to speckle noise (multiplicative Gaussian) and also weak in response to salt and pepper noise. This follows from the noise-sensitivity in morphological operations and should be taken into consideration when choosing an enhancement method.

% FIGURE
\renewcommand\sfigure{0.15}
\begin{figure*}[h!]
	\centering
	\begin{subfigure}[t]{\sfigure\linewidth}
	\includegraphics[width=\linewidth,height=\linewidth,keepaspectratio=true]{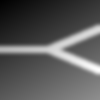}
	\caption{\quad}\label{fig:illum:orig}
	\end{subfigure}\hspace{10pt}
	\begin{subfigure}[t]{\sfigure\linewidth}
	\includegraphics[width=\linewidth,height=\linewidth,keepaspectratio=true]{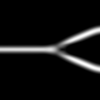}
	\caption{\quad}\label{fig:illum:vessel}
	\end{subfigure}
	\begin{subfigure}[t]{\sfigure\linewidth}
	\includegraphics[width=\linewidth,height=\linewidth,keepaspectratio=true]{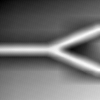}
	\caption{\quad}\label{fig:illum:clahe}
	\end{subfigure}
	\begin{subfigure}[t]{\sfigure\linewidth}
	\includegraphics[width=\linewidth,height=\linewidth,keepaspectratio=true]{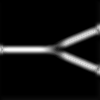}
	\caption{\quad}\label{fig:illum:soth}
	\end{subfigure}
	\begin{subfigure}[t]{\sfigure\linewidth}
	\includegraphics[width=\linewidth,height=\linewidth,keepaspectratio=true]{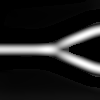}
	\caption{\quad}\label{fig:illum:neurite}
	\end{subfigure}
	\begin{subfigure}[t]{\sfigure\linewidth}
	\includegraphics[width=\linewidth,height=\linewidth,keepaspectratio=true]{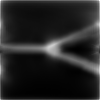}
	\caption{\quad}\label{fig:illum:pctv}
	\end{subfigure}
	\begin{subfigure}[t]{\sfigure\linewidth}
	\includegraphics[width=\linewidth,height=\linewidth,keepaspectratio=true]{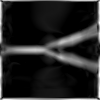}
	\caption{\quad}\label{fig:illum:pctn}
	\end{subfigure}
	\begin{subfigure}[t]{\sfigure\linewidth}
	\includegraphics[width=\linewidth,height=\linewidth,keepaspectratio=true]{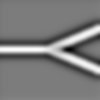}
	\caption{\quad}\label{fig:illum:wavelet}
	\end{subfigure}
	\begin{subfigure}[t]{\sfigure\linewidth}
	\includegraphics[width=\linewidth,height=\linewidth,keepaspectratio=true]{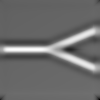}
	\caption{\quad}\label{fig:illum:bcosfire}
	\end{subfigure}
	\begin{subfigure}[t]{\sfigure\linewidth}
	\includegraphics[width=\linewidth,height=\linewidth,keepaspectratio=true]{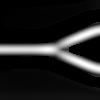}
	\caption{\quad}\label{fig:illum:vr}
	\end{subfigure}
	\begin{subfigure}[t]{\sfigure\linewidth}
	\includegraphics[width=\linewidth,height=\linewidth,keepaspectratio=true]{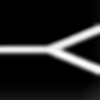}
	\caption{ours}\label{fig:illum:proposed}
	\end{subfigure}
	\hspace{-70pt}\vspace{-10pt}
	\caption{Comparison of the vessel enhancement methods' abilities to deal with an uneven background illumination. (\subref{fig:illum:orig})~an input image,
		(\subref{fig:illum:vessel})~vesselness,
		(\subref{fig:illum:clahe})~CLAHE,
		(\subref{fig:illum:soth})~Zana's top-hat,
		(\subref{fig:illum:neurite})~neuriteness,
		(\subref{fig:illum:pctv})~PCT vesselness,
		(\subref{fig:illum:pctn})~PCT neuriteness,
		(\subref{fig:illum:wavelet})~wavelet,
		(\subref{fig:illum:bcosfire})~line detector,
		(\subref{fig:illum:vr})~volume ratio, and (\subref{fig:illum:proposed})~the bowler-hat.}\label{fig:illum}
%\vspace{-10pt}
\end{figure*}

\subsection{Response to Uneven Background Illumination}\label{subsubsec:illum}

~\Cref{fig:illum} presents the response of the proposed method to an uneven illumination scenario. Key features such as junctions are preserved and appear unaffected by even severe illumination problems. This ability to preserve junctions under uneven illumination is important for many real applications of vessel enhancement and the proposed method is able to do this, unlike the current state-of-the-art methods. 

\subsection{Real Data - Retinal Image Datasets}\label{subsec:real}

In this section, we show the quality of the proposed method validated on three publicly available retinal image datasets: the DRIVE, STARE, and HRF databases. These datasets have been chosen because of their availability and their ground truth data. We have used these ground truth segmentations to quantitatively compare the proposed method with the other vessel enhancement methods.

The Digital Retinal Images for Vessel Extraction (DRIVE)~\cite{staal2004} dataset is a published database of retinal images for research and educational purposes. The database consists of twenty colour images that are JPEG compressed, as for many screening programs. These images were selected randomly from a screening of 400 diabetic subjects between the ages of 25 and 90. The ground truth provided with this dataset consists of manual segmentation of the vasculature for each image. Ground truths were prepared by trained observers, and 'true' pixels are those for which observers where $> 70\%$ certain.

The STructured Analysis of the REtina (STARE) dataset is another publicly available database~\cite{hoover2000locating} containing twenty colour images with human-determined vasculature ground truth. We have compared all these images against the AH labelling.

The High-Resolution Fundus (HRF) image dataset~\cite{odstrcilik2013retinal} consists of 45 retinal images. This dataset has three type of subjects include healthy, diabetic retinopathy, and glaucoma.\vspace{-10pt}

\subsubsection{Quantitative Validation - Enhancement}\label{subsec:quant}

While a visual inspection can give some information regarding the effectiveness of the vessel enhancement methods, a form of quantitative validation is required. Therefore, as proposed in~\cite{fawcett2006introduction}, we have used the Receiver Operating Characteristic (ROC) curve and we calculated the Area Under the Curve (AUC) based on ROC curve to compare the vessel enhancement methods. To derive the ROC curve and then to calculate the AUC value, each enhanced image is segmented at different thresholds and compared with the corresponding ground truth segmentation. Such a procedure is used for~\Cref{fig:psnr},~\Cref{fig:roc}, and~\Cref{tab:auc}.

\subsubsection{Quantitative Validation - Segmentation}\label{subsec:quant:seg}

To quantitatively evaluate the robustness of the vessel segmentation methods, sensitivity (SE), specificity (SP), and accuracy (ACC) metrics are calculated for each segmented image and its corresponding ground truth segmentation, as follows: 

	\begin{align}
		&\text{SE} = \frac{TP}{TP+FN},\\
		&\text{SP} = \frac{TN}{TN+FP},\\
		&\text{ACC} = \frac{TP+TN}{TP+TN+FP+FN},
\end{align}
where $TP$ is the true positive count, $FP$ the false positive count, $TN$ the true negative and $FN$ the false negative counts of the segmented pixels. We used these metrics in~\Cref{tab:acc} and~\Cref{tab:seg}.

\subsubsection{Healthy Subjects}\label{subsubsec:real:compare}

~\Cref{fig:hrf} shows the results of the proposed and state-of-the-art methods applied to a sample image from the HRF dataset (results for DRIVE and STARE datasets can be found in Supplementary Materials).

We can see that the proposed method is able to enhance finer structures as detected by the human observer but not emphasised by many of the other methods (see arrows). 

We can also see that, whilst the connectivity seems to be maintained (unlike in~\Cref{fig:hrf:vessel}), `false vessels' are not introduced (\cf{}~\Cref{fig:hrf:neurite}).

Finally,~\Cref{fig:roc} and~\Cref{tab:auc} present ROC curves and mean AUC values for the enhancement results of the proposed and state-of-the-art methods applied to all images across the DRIVE, STARE and HRF datasets by using the quantitative validation as described~\Cref{subsec:quant}. 

% FIGURE
\renewcommand\sfigure{0.31}
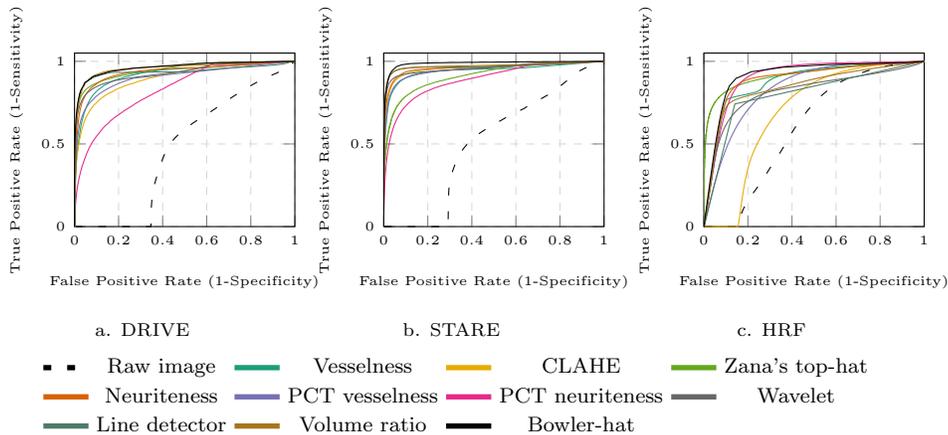
\begin{figure}[ht!]
	\begin{subfigure}{\sfigure\linewidth}
		\begin{tikzpicture}
		\begin{axis}[
		width = 1.2\linewidth,
		xlabel={False Positive Rate (1-Specificity)},
		ylabel={True Positive Rate (1-Sensitivity)},
		ylabel near ticks,
		label style={font=\tiny},
		tick label style={font=\tiny},
		ytickmin=0, ymax=1.05,
		xtickmin=0, xtickmax=1.05,
		enlargelimits=false,
		legend entries = {Raw image, Vesselness, CLAHE,Zana's top-hat, Neuriteness, PCT vesselness, PCT neuriteness, Wavelet, Line detector, Volume ratio, Bowler-hat},
		legend columns = 4,
		legend style={font=\footnotesize,line width=2pt, draw=none,},% cells={anchor=east}},
		legend to name=leg3,
		grid=major, % Display a grid
		grid style={dashed,gray!30}, % Set the style
		]
		\addplot[color=black,loosely dashed] table [x index=0, y index=1, col sep=comma] {image/datFiles/driveO.dat};%Raw Image
		\addplot[color=brewerDark1] table [x index=0, y index=1, col sep=comma] {image/datFiles/driveV.dat};%Vesselness
		\addplot[color=brewerDark6] table [x index=0, y index=1, col sep=comma] {image/datFiles/driveClahe.dat};%Clahe
		\addplot[color=brewerDark5] table [x index=0, y index=1, col sep=comma] {image/datFiles/driveSsum.dat};%Top-hats
		\addplot[color=brewerDark2] table [x index=0, y index=1, col sep=comma] {image/datFiles/driveN.dat};%Neuriteness
		\addplot[color=brewerDark3] table [x index=0, y index=1, col sep=comma] {image/datFiles/drivePV.dat};%Pct Vess
		\addplot[color=brewerDark4] table [x index=0, y index=1, col sep=comma] {image/datFiles/drivePN.dat};%Pct Neurite
		\addplot[color=brewerDark8] table [x index=0, y index=1, col sep=comma] {image/datFiles/driveWa.dat};%Wavelet
		\addplot[color=green!70] table [x index=0, y index=1, col sep=comma] {image/datFiles/driveB.dat};%line detector
		\addplot[color=brewerDark7] table [x index=0, y index=1, col sep=comma] {image/datFiles/driveVR.dat};%Volume Ratio
		\addplot[color=black] table [x index=0, y index=1, col sep=comma] {image/datFiles/driveP.dat};%proposed
		\end{axis}
		\end{tikzpicture}
		\caption{DRIVE}\label{fig:roc:drive}
	\end{subfigure}\quad
	\begin{subfigure}{\sfigure\linewidth}
		\begin{tikzpicture}
		\begin{axis}[
		width = 1.2\linewidth,
		xlabel={False Positive Rate (1-Specificity)},
		ylabel={True Positive Rate (1-Sensitivity)},
		ylabel near ticks,
		label style={font=\tiny},
		tick label style={font=\tiny},
		ytickmin=0, ymax=1.05,
		xtickmin=0, xtickmax=1.05,
		enlargelimits=false,
		grid=major, % Display a grid
		grid style={dashed,gray!30}, % Set the style
		]
		\addplot[color=black,loosely dashed] table [x index=0, y index=1, col sep=comma] {image/datFiles/stareO.dat};%Raw Image
		\addplot[color=brewerDark1] table [x index=0, y index=1, col sep=comma,header=false] {image/datFiles/stareV.dat};%Vesselness
		\addplot[color=brewerDark6] table [x index=0, y index=1, col sep=comma,header=false] {image/datFiles/stareClahe.dat};%Clahe
		\addplot[color=brewerDark5] table [x index=0, y index=1, col sep=comma,header=false] {image/datFiles/stareSsum.dat};%Top-hats
		\addplot[color=brewerDark2] table [x index=0, y index=1, col sep=comma,header=false] {image/datFiles/stareN.dat};%Neuriteness
		\addplot[color=brewerDark3] table [x index=0, y index=1, col sep=comma,header=false] {image/datFiles/starePV.dat};%Pct Vess
		\addplot[color=brewerDark4] table [x index=0, y index=1, col sep=comma,header=false] {image/datFiles/starePN.dat};%Pct Neurite
		\addplot[color=brewerDark8] table [x index=0, y index=1, col sep=comma,header=false] {image/datFiles/stareWa.dat};%Wavelet
		\addplot[color=green!70] table [x index=0, y index=1, col sep=comma,header=false] {image/datFiles/stareB.dat};%line detector
		\addplot[color=brewerDark7] table [x index=0, y index=1, col sep=comma,header=false] {image/datFiles/stareVR.dat};%Vr
		\addplot[color=black] table [x index=0, y index=1, col sep=comma,header=false] {image/datFiles/stareP.dat};%proposed
		\end{axis}
		\end{tikzpicture}
		\caption{STARE}\label{fig:roc:stare}
	\end{subfigure}
	\begin{subfigure}{\sfigure\linewidth}
		\begin{tikzpicture}
		\begin{axis}[
		width = 1.2\linewidth,
		xlabel={False Positive Rate (1-Specificity)},
		ylabel={True Positive Rate (1-Sensitivity)},
		ylabel near ticks,
		label style={font=\tiny},
		tick label style={font=\tiny},
		ytickmin=0, ymax=1.05,
		xtickmin=0, xtickmax=1.05,
		enlargelimits=false,
		grid=major, % Display a grid
		grid style={dashed,gray!30},]
		\addplot[color=black,loosely dashed] table [x index=0, y index=1, col sep=comma] 	{image/datFiles/hrfO.dat};%Raw Image
		\addplot[color=brewerDark1] table [x index=0, y index=1, col sep=comma,header=false] 	{image/datFiles/hrfV.dat};%Vesselness
		\addplot[color=brewerDark6] table [x index=0, y index=1, col sep=comma,header=false] 	{image/datFiles/hrfClahe.dat};%Clahe
		\addplot[color=brewerDark5] table [x index=0, y index=1, col sep=comma,header=false] 	{image/datFiles/hrfSsum.dat}; %Top-hat
		\addplot[color=brewerDark2] table [x index=0, y index=1, col sep=comma,header=false] 	{image/datFiles/hrfN.dat};%Neuriteness
		\addplot[color=brewerDark3] table [x index=0, y index=1, col sep=comma,header=false] 	{image/datFiles/hrfPV.dat};%Pct Vess
		\addplot[color=brewerDark4] table [x index=0, y index=1, col sep=comma,header=false] 	{image/datFiles/hrfPN.dat};%Pct Neurite
		\addplot[color=brewerDark8] table [x index=0, y index=1, col sep=comma,header=false] {image/datFiles/hrfWa.dat};%Wavelet
		\addplot[color=green!70] table [x index=0, y index=1, col sep=comma,header=false] {image/datFiles/hrfB.dat};%line detector
		\addplot[color=brewerDark7] table [x index=0, y index=1, col sep=comma,header=false] 	{image/datFiles/hrfVR.dat};%Vr
		\addplot[color=black] table [x index=0, y index=1, col sep=comma,header=false] {image/datFiles/hrfP.dat};
		\end{axis}
		\end{tikzpicture}
		\caption{HRF}\label{fig:roc:hrfhealthy}
	\end{subfigure}
	\begin{subfigure}{\linewidth}
		\centering
		\pgfplotslegendfromname{leg3}
	\end{subfigure}\vskip-4pt
	\caption{ROC curves calculated for sample images from the (\subref{fig:roc:drive}) DRIVE, (\subref{fig:roc:stare}) STARE, and (\subref{fig:roc:hrfhealthy}) HRF datasets enhanced by the proposed and the state-of-the-art methods (see legend for colours). Corresponding mean AUC values can be found in~\Cref{tab:auc}.}\label{fig:roc}
\end{figure}

\subsubsection{Unhealthy Subjects}

~\Cref{fig:unhealthy} presents a visual comparison of the enhancement methods applied to sample images of subjects with diabetic retinopathy and with glaucoma from the DRIVE, STARE and HRF datasets. As we can notice in~\Cref{fig:pathology:stareResult}, the proposed method is sensitive to noisy regions. This issue can be addressed by the use of a line-shaped morphological structuring element with a varying thickness. 
	Even so, the proposed method achieved a highest overall score on the HRF unhealthy images as illustrated in~\Cref{tab:auc}.\vspace{-10pt}

%FIGURE
\input{./tex/figure-hrf}
% TABLE
\makeatletter
\newcommand{\thickhline}{%
	\noalign {\ifnum 0=`}\fi \hrule height 1pt
	\futurelet \reserved@a \@xhline
}
\begin{table}[h!]
	\resizebox{\columnwidth}{!}{%
	\centering
	\begin{tabular}{llcccc}
		\toprule
		\multirow{2}{2.5cm}{\centering Enhancement Method}&\multicolumn{4}{c}{AUC (StDev)}\\ \cmidrule(l){2-6}\cmidrule(l){2-6}
		 &Year/Ref\quad &DRIVE\quad&STARE\quad&HRF(healthy)\quad&HRF(unhealthy)\quad\\
		\midrule
		Raw	image		&-									&0.416 (0.064) 			&0.490 (0.076)			&0.530 (0.075)				&0.541(0.073)\\
		\midrule
		Vesselness		&1998~\cite{frangi1998multiscale}	&0.888 (0.243) 			&0.898 (0.215)			&0.913 (0.020)				&0.904 (0.020)\\
		\midrule
		CLAHE			&1998~\cite{Pisano1998}				&0.862 (0.068) 			&0.880 (0.087)			&0.867 (0.025)				&0.835 (0.023)\\
		\midrule
		Zana's top-hat 	&2001~\cite{ZK2001}					&0.933 (0.015) 			&0.956 (0.021)			&0.943 (0.010)				&0.91 (0.016)\\
		\midrule
		Neuriteness		&2004~\cite{meijering2004design}	&0.909 (0.022) 			&0.927 (0.039)			&0.896 (0.024)				&0.879 (0.059)\\
		\midrule
		PCT vesselness 	&2012~\cite{obara2012contrast}		&0.890 (0.037) 			&0.899 (0.056)			&0.888 (0.011)				&0.837 (0.030)\\
		\midrule
		PCT neuriteness	&2012~\cite{obara2012contrast} 		&0.817 (0.121) 			&0.827 (0.165)			&0.901 (0.029)				&0.777 (0.022)\\
		\midrule
		Wavelet			&2012~\cite{bankhead2012fast}	 	&0.891 (0.024) 			&0.867 (0.042)			&0.802 (0.022)				&0.740 (0.026)\\
		\midrule
		Line detector	&2013~\cite{nguyen2013effective}	&0.828 (0.024) 			&0.856 (0.042)			&0.820 (0.022)				&0.734 (0.026)\\
		\midrule
		Volume ratio	&2016~\cite{JPLetal2016} 			&0.934 (0.024) 			&0.939 (0.042)			&0.926 (0.022)				&0.823 (0.026)\\
		\midrule
		Bowler-hat		&-									&\textbf{0.946 (0.032)}	&\textbf{0.962 (0.034)}	&\textbf{0.968 (0.015)} 	&\textbf{0.944 (0.016)}\\
		\thickhline\thickhline
	\end{tabular}
}
	\caption{Mean AUC values calculated as described in~\Cref{subsec:quant}, for the images across the DRIVE, STARE and HRF datasets enhanced by the bowler-hat and the state-of-the-art methods. Best results for each dataset are in bold. Individual ROC curves can be seen in~\Cref{fig:roc}.}\label{tab:auc}
\end{table}

\subsubsection{Enhancement with Local Thresholding}\label{subsubsec:seg}

~\Cref{fig:seg} and~\Cref{tab:acc} demonstrate the vessel segmentation results obtained by the proposed and the state-of-the-art vessel-like structures enhancement methods followed by the same local thresholding approach proposed in~\cite{jiang2003adaptive} when applied to the HRF dataset images.

%FIGURE
\renewcommand\sfigure{0.3}
\begin{figure}[hb!]
	\centering
	%DRIVE
	\begin{subfigure}[t]{0.01\linewidth}
		\centering
	\end{subfigure}
	\begin{subfigure}[t]{\sfigure\linewidth}
		\begin{tikzpicture}[scale=1]
		\node(0,0){\includegraphics[trim={20 20 20 20}, clip,width=2.75cm,height=2.5cm]{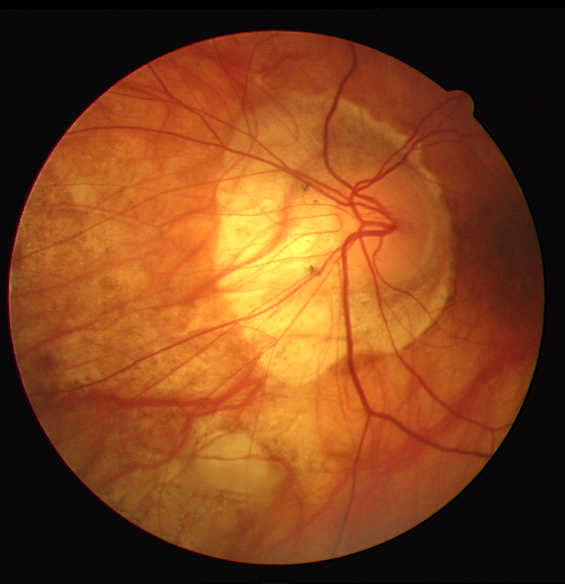}};
		\draw[line width=1pt, white] (-.3,-.3)rectangle(.8,.7) node[pos=0.5] {\textbf{1}};
		\end{tikzpicture}
		\vskip-0.3cm
		\caption{\quad}\label{fig:pathology:driveRoI}
	\end{subfigure}
\hspace{-1cm}
	\begin{subfigure}[t]{\sfigure\linewidth}
		\begin{tikzpicture}[scale=1]
		\node(0,0){\includegraphics[trim={40 0 40 0}, clip,width=2.75cm,height=2.5cm]{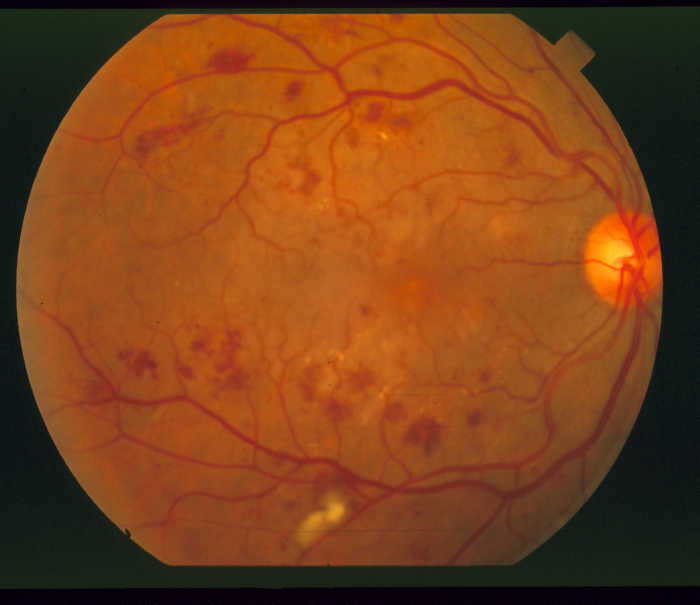}};
		\draw[line width=1pt, white] (-.9,.1)rectangle(.1,1.1) node[pos=0.5] {\textbf{2}};
		\end{tikzpicture}
		\vskip-0.3cm
		\caption{\quad}\label{fig:pathology:stareRoI}
	\end{subfigure}
\hspace{-1cm}
	\begin{subfigure}[t]{\sfigure\linewidth}
		\begin{tikzpicture}[scale=1]
		\node(0,0){\includegraphics[trim={30 0 30 0}, clip,width=2.75cm,height=2.5cm]{hrf}};
		\draw[line width=1pt, white] (-.2,.2)rectangle(.7,1.15) node[pos=0.5] {\textbf{3}};
		\end{tikzpicture}
		\vskip-0.3cm
		\caption{\quad}\label{fig:pathology:hrfRoI}
	\end{subfigure}\\ 
	\begin{subfigure}[t]{0.01\linewidth}
		\centering
		\vskip-40pt
		\textbf{1}
	\end{subfigure}
	\begin{subfigure}[t]{\sfigure\linewidth}
		\begin{tikzpicture}[scale=1]
		\node(0,0){\includegraphics[trim={190 210 130 135}, clip,width=2.75cm,height=2.75cm,,keepaspectratio=true]{drive}};
		\end{tikzpicture}
		\vskip-0.3cm
		\caption{\quad}\label{fig:pathology:drive}
	\end{subfigure}
\hspace{-1cm}
	\begin{subfigure}[t]{\sfigure\linewidth}
		\begin{tikzpicture}[scale=1]
		\node(0,0){\includegraphics[trim={190 210 130 135}
			, clip,width=2.75cm,height=2.75cm,keepaspectratio=true]{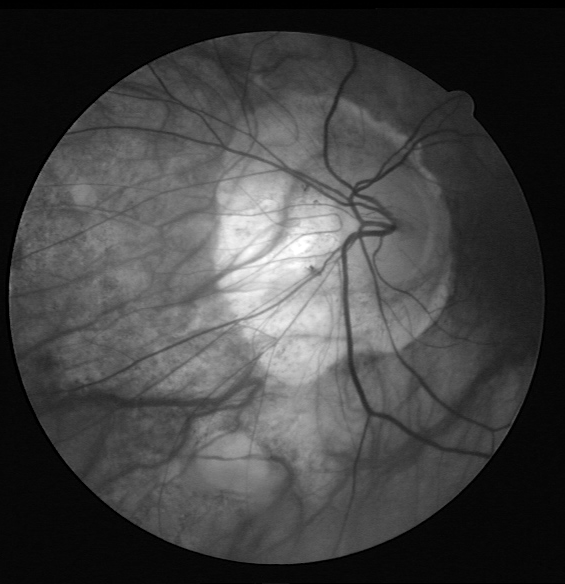}};
		\end{tikzpicture}
		\vskip-0.3cm
		\caption{\quad}\label{fig:pathology:driveGray}
	\end{subfigure}
\hspace{-1cm}
	\begin{subfigure}[t]{\sfigure\linewidth}
		\begin{tikzpicture}[scale=1]
		\node(0,0){\includegraphics[trim={190 210 130 135}, clip,width=2.75cm,height=2.75cm,keepaspectratio=true]{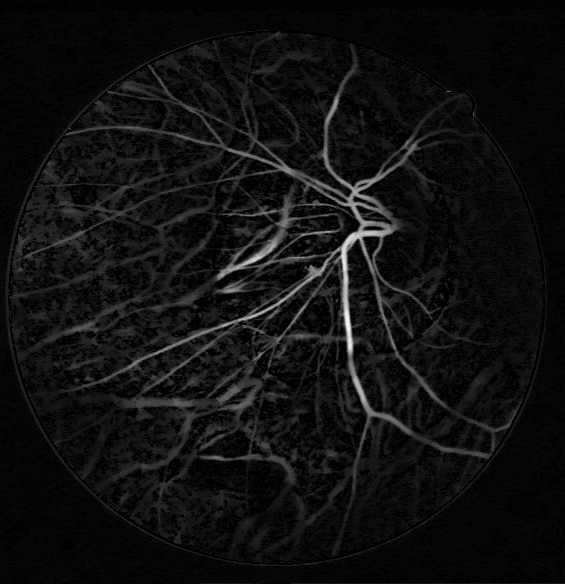}};
		\end{tikzpicture}
		\vskip-0.3cm
		\caption{\quad}\label{fig:pathology:driveResult}
	\end{subfigure}\\
	\begin{subfigure}[t]{0.01\linewidth}
		\centering
		\vskip-40pt
		\textbf{2}
	\end{subfigure}
	%STARE
	\begin{subfigure}[t]{\sfigure\linewidth}
		\begin{tikzpicture}[scale=1]
		\node(0,0){	\includegraphics[trim={150 325 310 50}, clip,width=2.75cm,height=3.25cm,keepaspectratio=true]{stare}};
		\end{tikzpicture}
		\vskip-0.3cm
		\caption{\quad}\label{fig:pathology:stare}
	\end{subfigure}
\hspace{-1cm}
	\begin{subfigure}[t]{\sfigure\linewidth}
		\begin{tikzpicture}[scale=1]
		\node(0,0){	\includegraphics[trim={150 325 310 50}
			, clip,width=2.75cm,height=3.25cm,keepaspectratio=true]{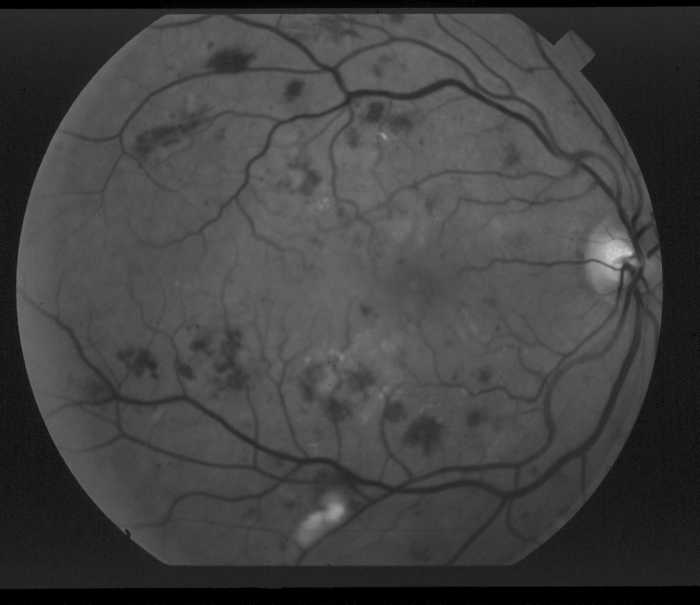}};
		\end{tikzpicture}
		\vskip-0.3cm
		\caption{\quad}\label{fig:pathology:stareGray}
	\end{subfigure}
\hspace{-1cm}
	\begin{subfigure}[t]{\sfigure\linewidth}
		\begin{tikzpicture}[scale=1]
		\node(0,0){	\includegraphics[trim={150 325 310 50}, clip,width=2.75cm,height=3.25cm,keepaspectratio=true]{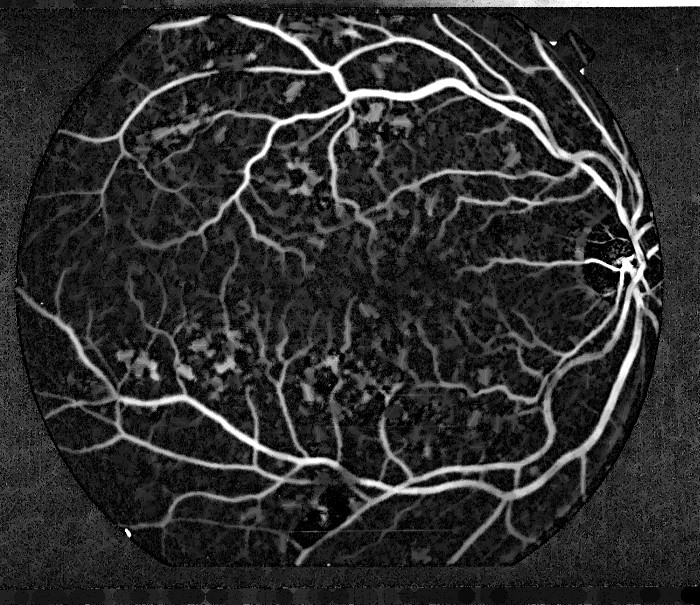}};
		\end{tikzpicture}
		\vskip-0.3cm
		\caption{\quad}\label{fig:pathology:stareResult}
	\end{subfigure}\\
	\begin{subfigure}[t]{0.01\linewidth}
		\centering
		\vskip-40pt
		\textbf{3}
	\end{subfigure}
	%HRF
	\begin{subfigure}[t]{\sfigure\linewidth}
		\begin{tikzpicture}[scale=1]
		\node(0,0){	\includegraphics[trim={380 340 240 10}, clip,width=2.75cm,height=3cm,keepaspectratio=true]{hrf}};
		\end{tikzpicture}
		\vskip-0.3cm
		\caption{\quad}\label{fig:pathology:hrf}
	\end{subfigure}
\hspace{-1cm}
	\begin{subfigure}[t]{\sfigure\linewidth}
		\begin{tikzpicture}[scale=1]
		\node(0,0){	\includegraphics[trim={280 251 180 6}, clip,width=2.75cm,height=3cm,keepaspectratio=true]{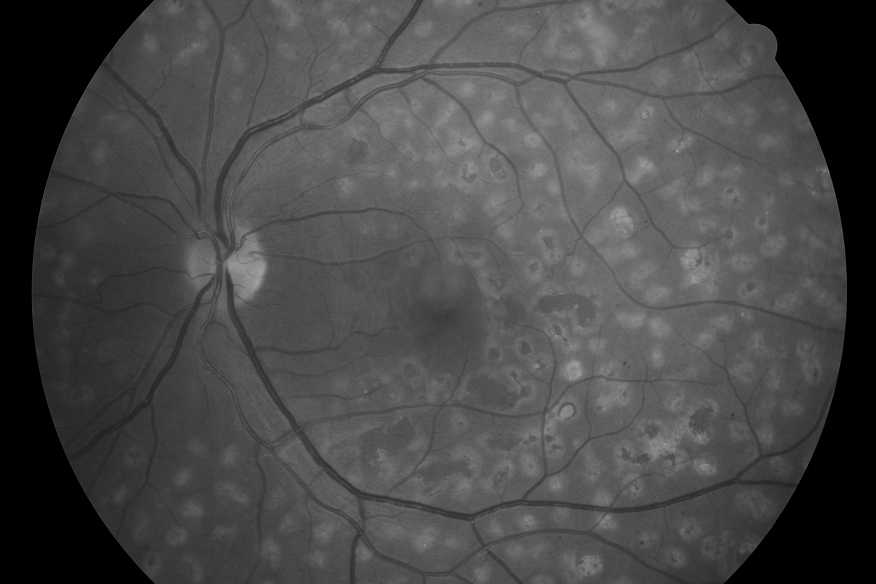}};
		%		\draw[line width=0.5pt, red!60] (0,-.6) rectangle (.6,-.1) node[pos=0.5] {};
		\end{tikzpicture}\label{fig:pathology:hrfGray}
		\vskip-0.3cm
		\caption{\quad}
	\end{subfigure}
\hspace{-1cm}
	\begin{subfigure}[t]{\sfigure\linewidth}
		\begin{tikzpicture}[scale=1]
		\node(0,0){	\includegraphics[trim={430 380 263 10}, clip,width=2.75cm,height=3cm,keepaspectratio=true]{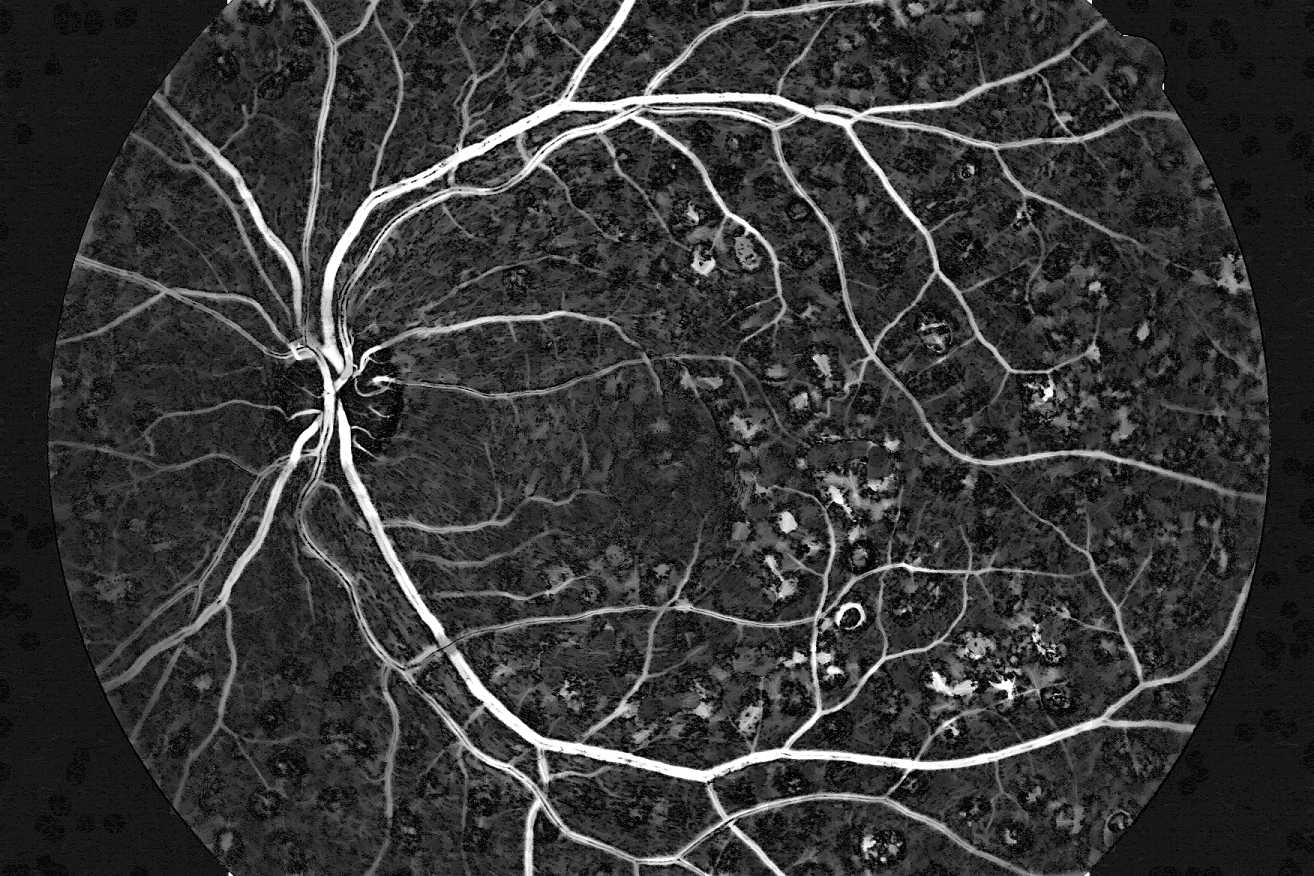}};
		\end{tikzpicture}
		\vskip-0.3cm
		\caption{\quad}\label{fig:pathology:hrfResult}
	\end{subfigure}
	\caption{The bowler-hat applied to the unhealthy subjects from
		(\subref{fig:pathology:driveRoI})~DRIVE,
		(\subref{fig:pathology:stareRoI})~STARE and
		(\subref{fig:pathology:hrfRoI})~HRF. 
		(\subref{fig:pathology:drive},
		\subref{fig:pathology:stare},
		\subref{fig:pathology:hrf}) are the input images with the region of interest. 
		(\subref{fig:pathology:driveGray},
		\subref{fig:pathology:stareGray},
		\subref{fig:pathology:hrfGray}) illustrate the green channel of input image % result after proposed method performed.
		(\subref{fig:pathology:driveResult},
		\subref{fig:pathology:stareResult},
		\subref{fig:pathology:hrfResult}) demonstrate the enhancement result of the vessel-like structure on the abnormal area.}\label{fig:unhealthy}
\end{figure}
% FIGURE
% !TEX root = ../main-elsevier.tex
\renewcommand\sfigure{0.19}
\begin{figure*}[h!]
	\centering
	\begin{subfigure}[t]{\sfigure\linewidth}
	\includegraphics[width=\linewidth,height=\linewidth,keepaspectratio=true]{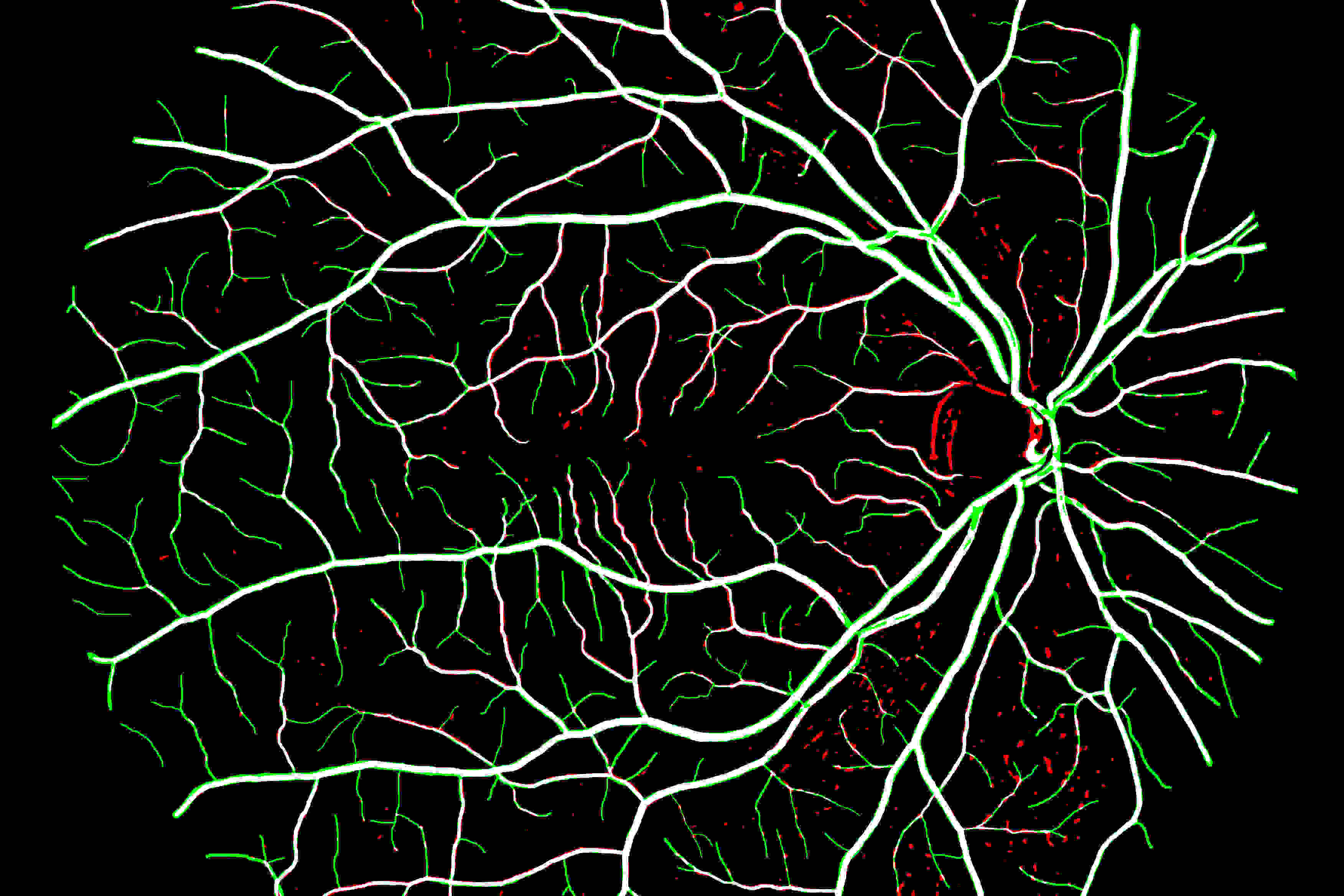}
	\caption{\quad}\label{fig:seg:vessel}
	\end{subfigure}
	\begin{subfigure}[t]{\sfigure\linewidth}
	\includegraphics[width=\linewidth,height=\linewidth,keepaspectratio=true]{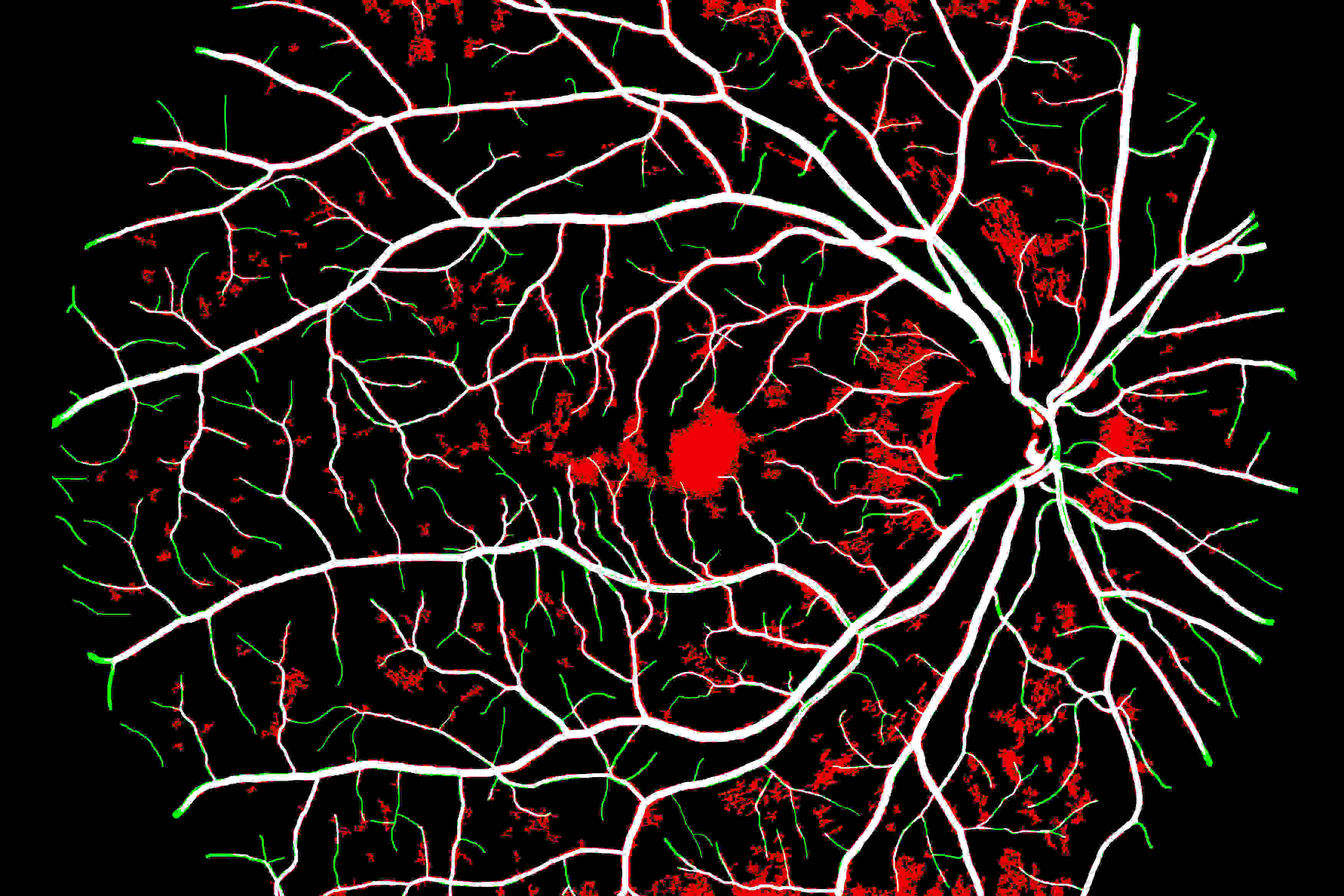}
	\caption{\quad}\label{fig:seg:clahe}
	\end{subfigure}
	\begin{subfigure}[t]{\sfigure\linewidth}
	\includegraphics[width=\linewidth,height=\linewidth,keepaspectratio=true]{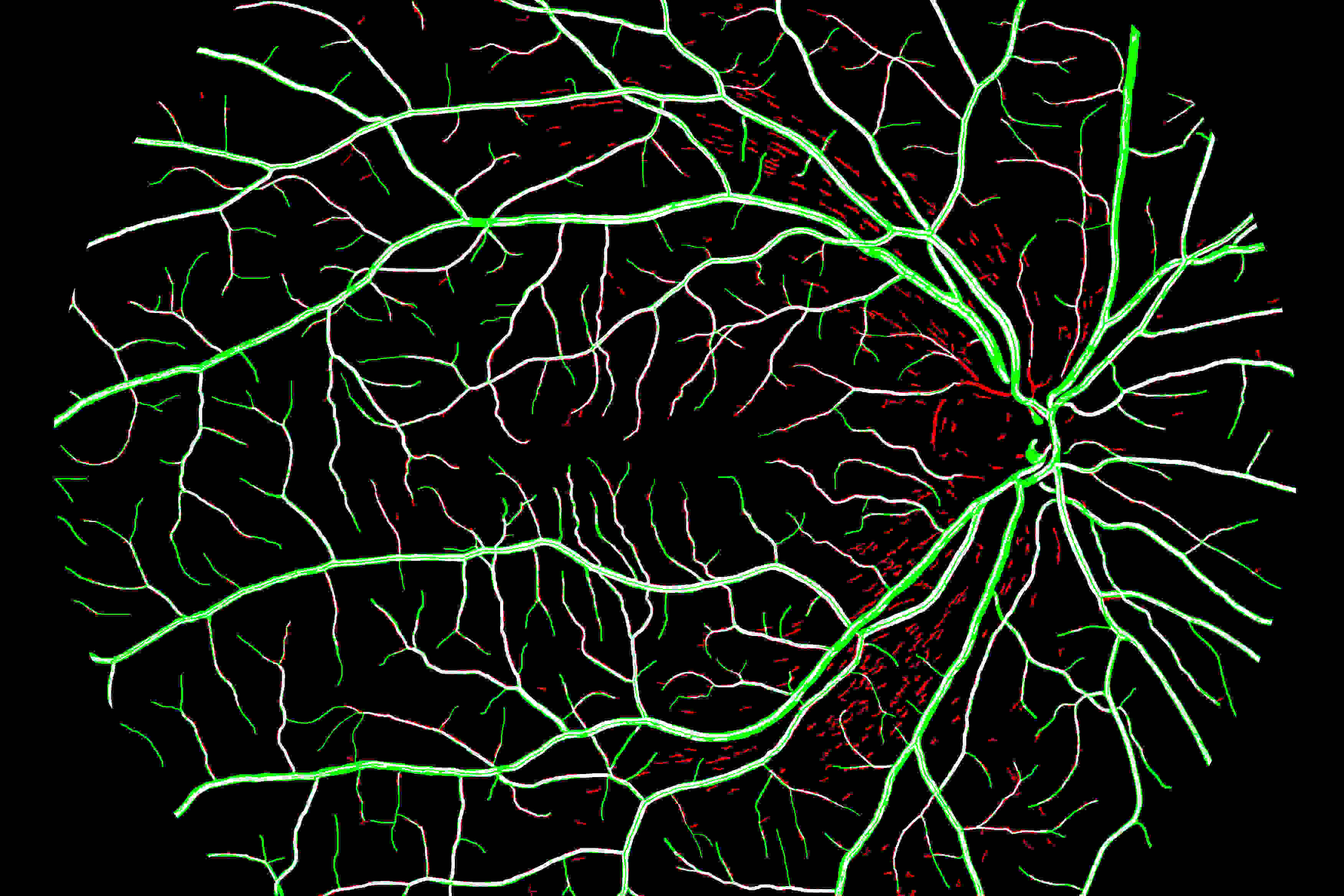}
	\caption{\quad}\label{fig:seg:soth}
	\end{subfigure}
	\begin{subfigure}[t]{\sfigure\linewidth}
	\includegraphics[width=\linewidth,height=\linewidth,keepaspectratio=true]{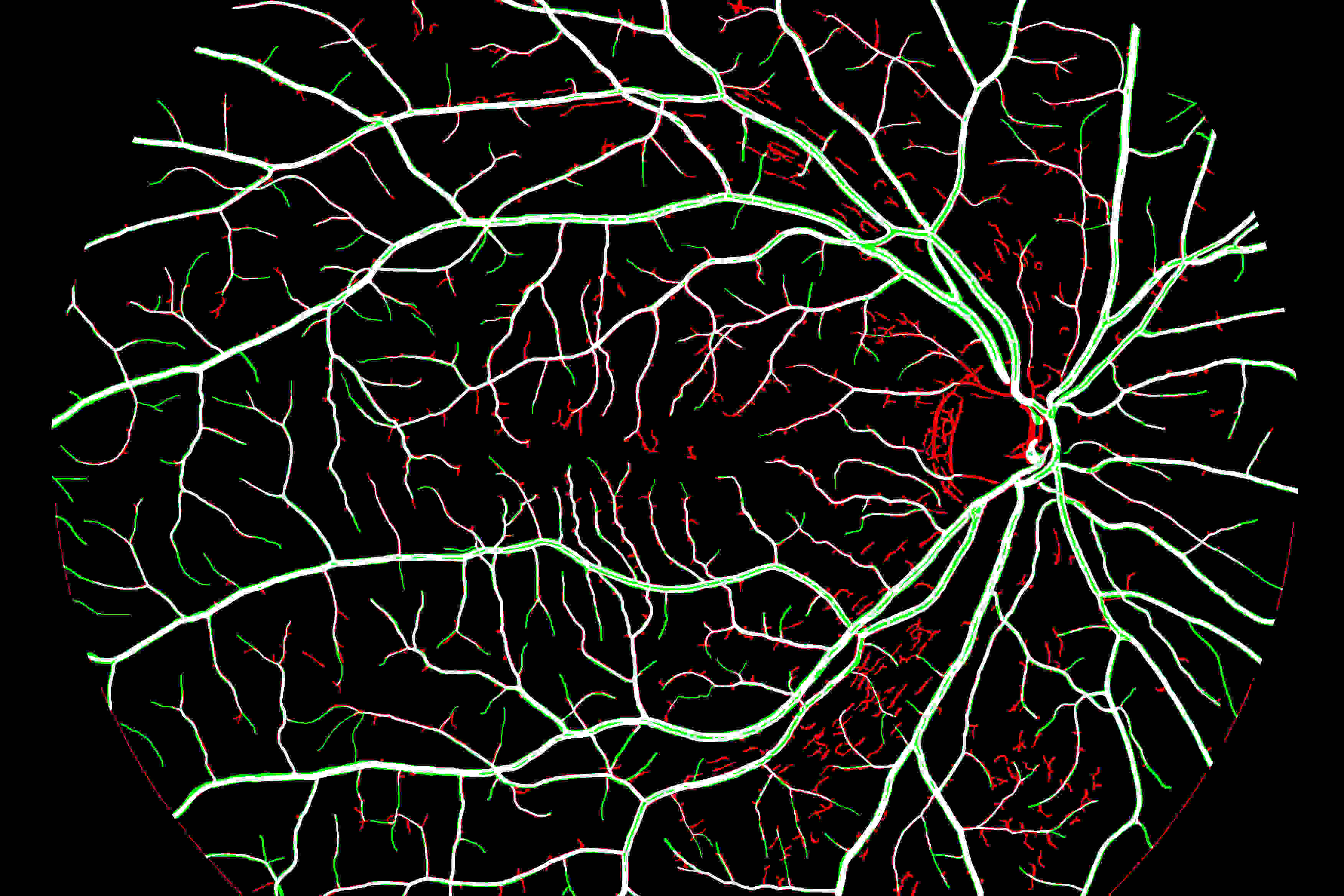}
	\caption{\quad}\label{fig:seg:neurite}
	\end{subfigure}
	\begin{subfigure}[t]{\sfigure\linewidth}
	\includegraphics[width=\linewidth,height=\linewidth,keepaspectratio=true]{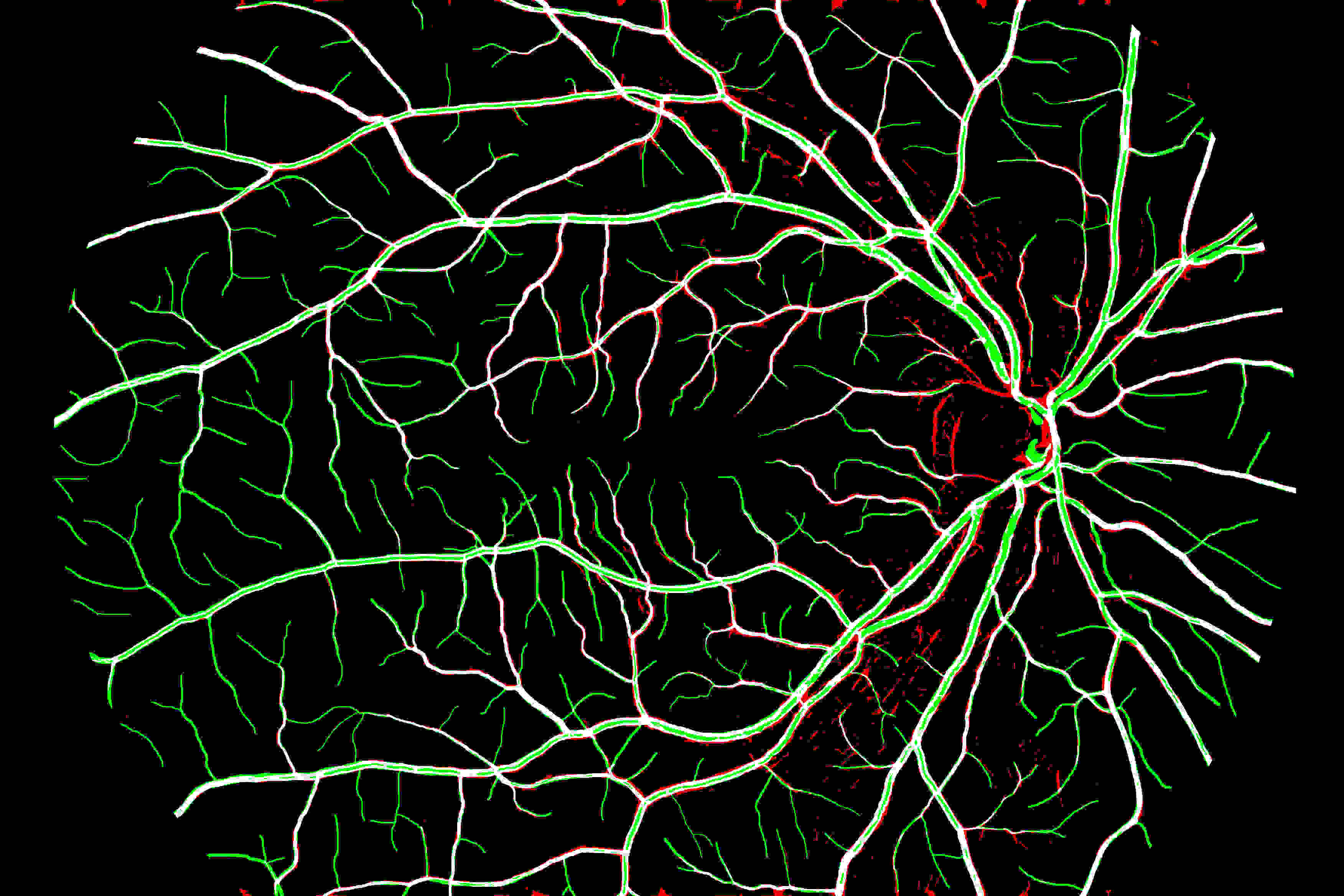}
	\caption{\quad}\label{fig:seg:pctv}
	\end{subfigure}
	\begin{subfigure}[t]{\sfigure\linewidth}
	\includegraphics[width=\linewidth,height=\linewidth,keepaspectratio=true]{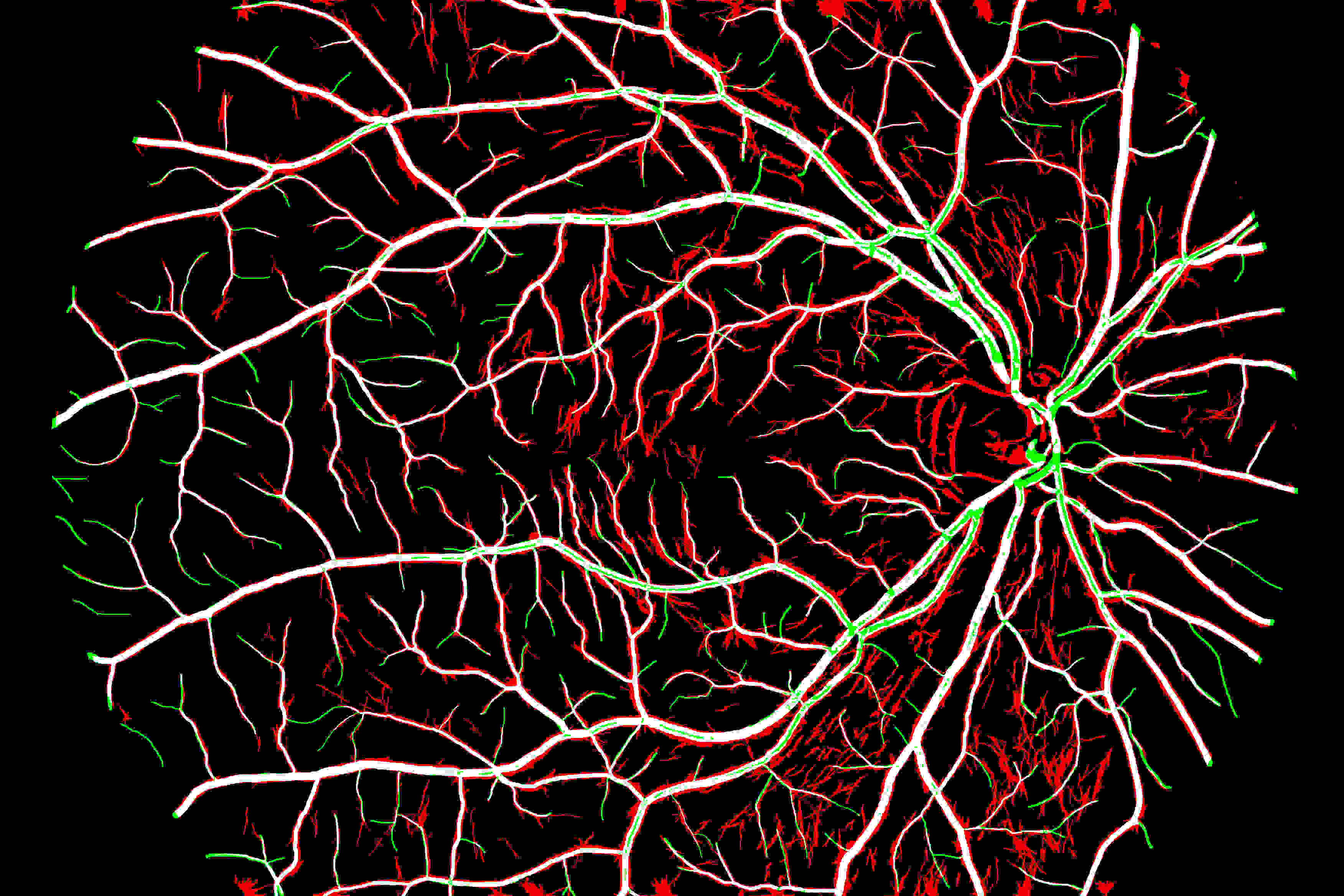}
	\caption{\quad}\label{fig:seg:pctn}
	\end{subfigure}
	\begin{subfigure}[t]{\sfigure\linewidth}
	\includegraphics[width=\linewidth,height=\linewidth,keepaspectratio=true]{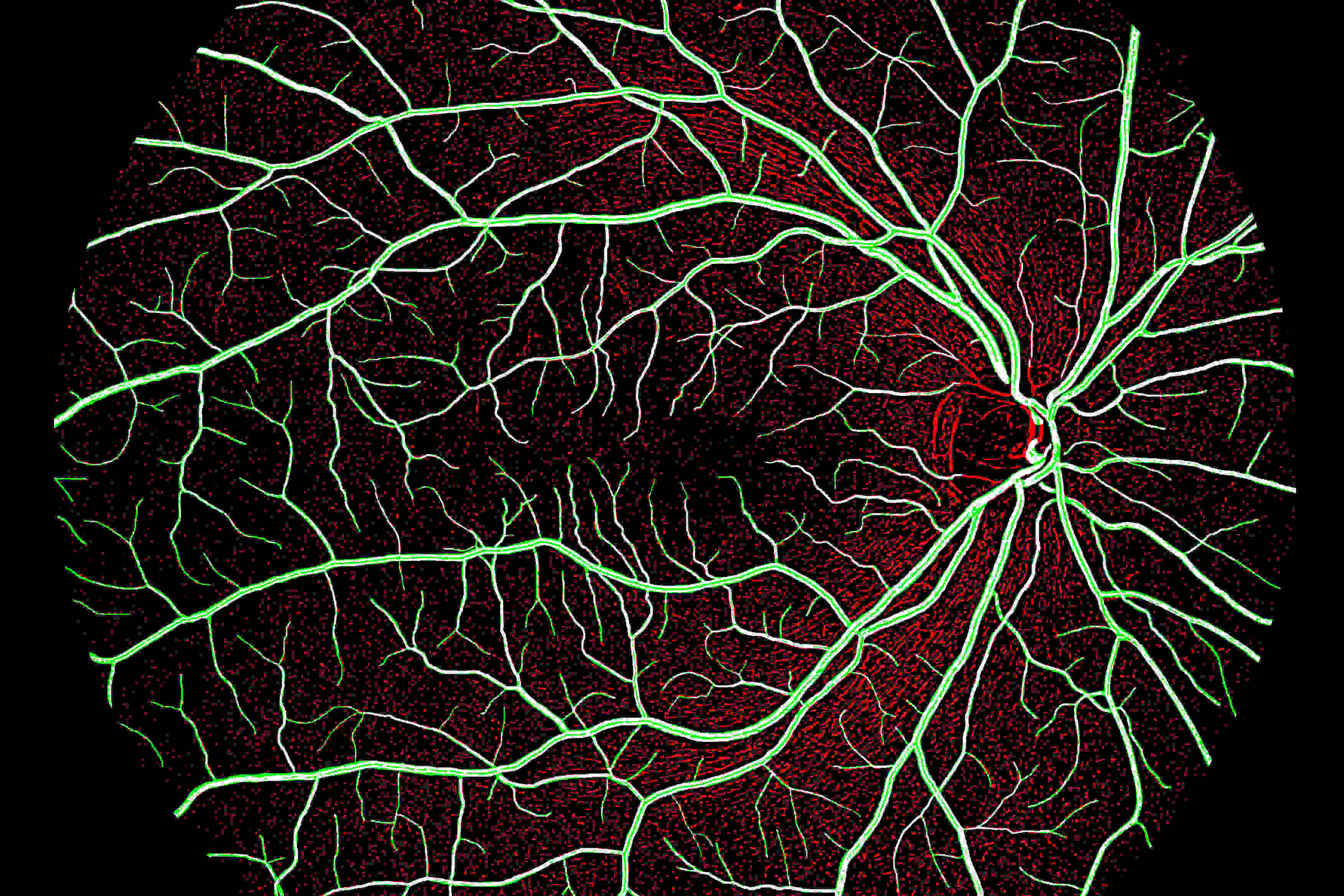}
	\caption{\quad}\label{fig:seg:wavelet}
	\end{subfigure}
	\begin{subfigure}[t]{\sfigure\linewidth}
	\includegraphics[width=\linewidth,height=\linewidth,keepaspectratio=true]{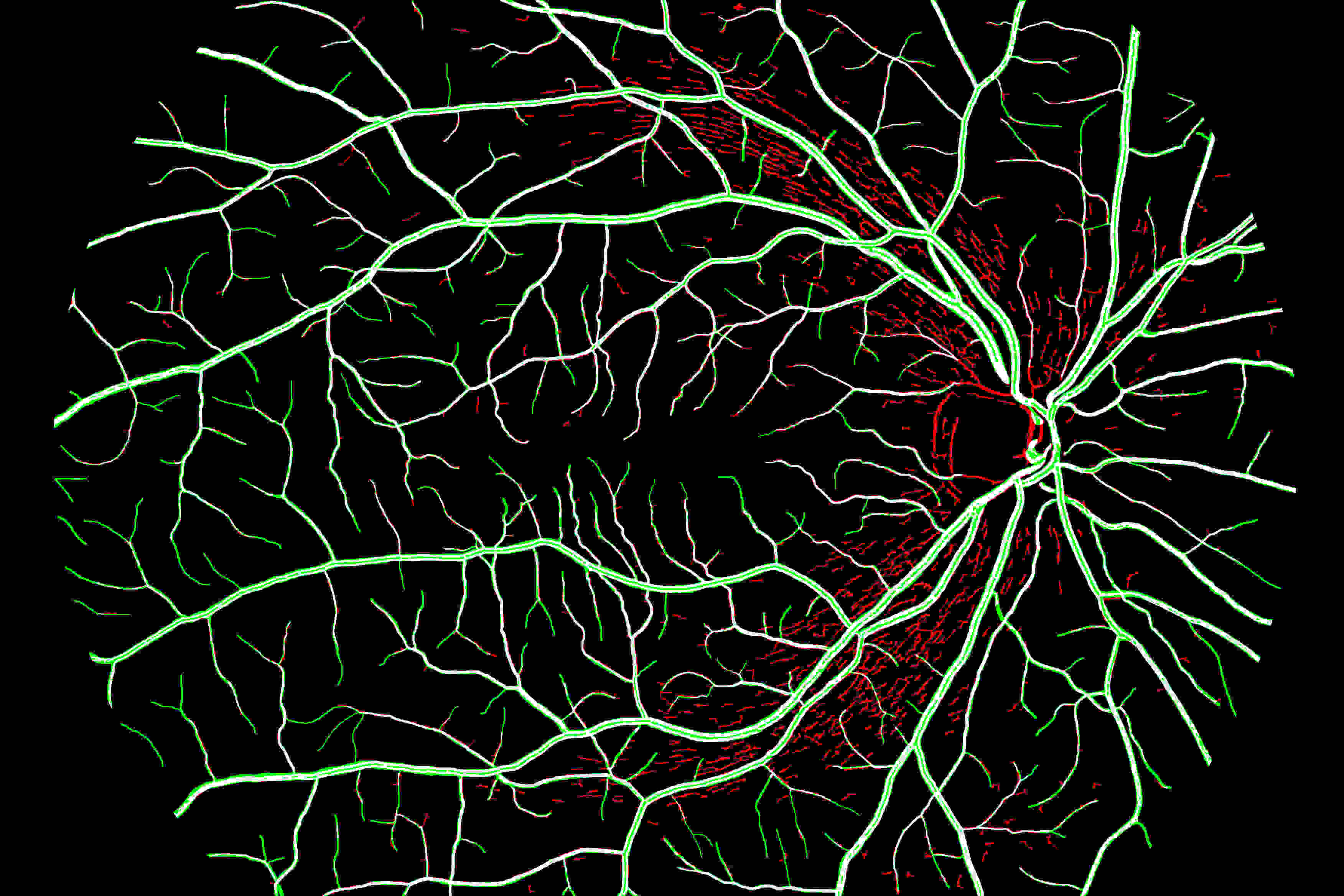}
	\caption{\quad}\label{fig:seg:bcosfire}
	\end{subfigure}
	\begin{subfigure}[t]{\sfigure\linewidth}
	\includegraphics[width=\linewidth,height=\linewidth,keepaspectratio=true]{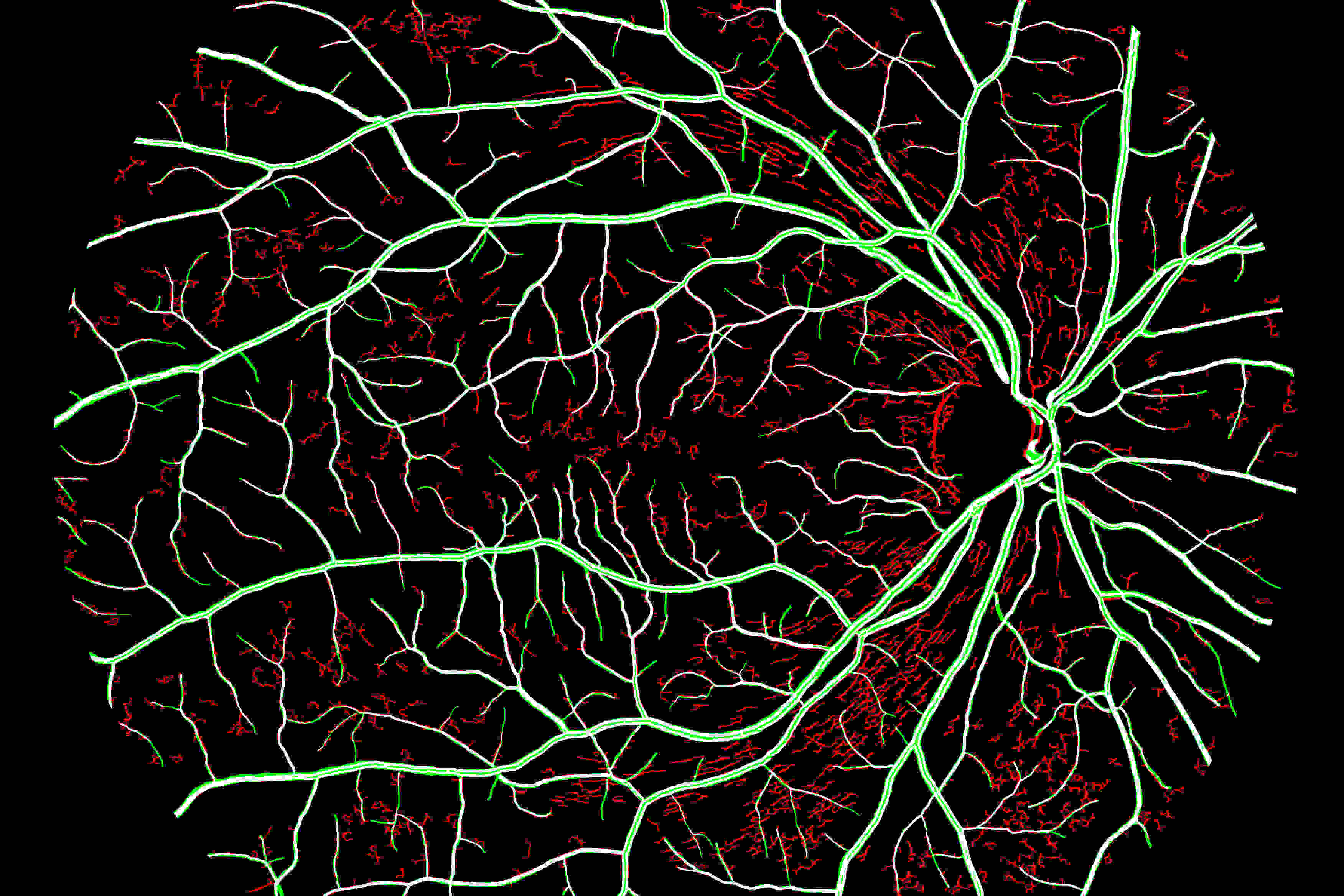}
	\caption{\quad}\label{fig:seg:vr}
	\end{subfigure}
	\begin{subfigure}[t]{\sfigure\linewidth}
	\includegraphics[width=\linewidth,height=\linewidth,keepaspectratio=true]{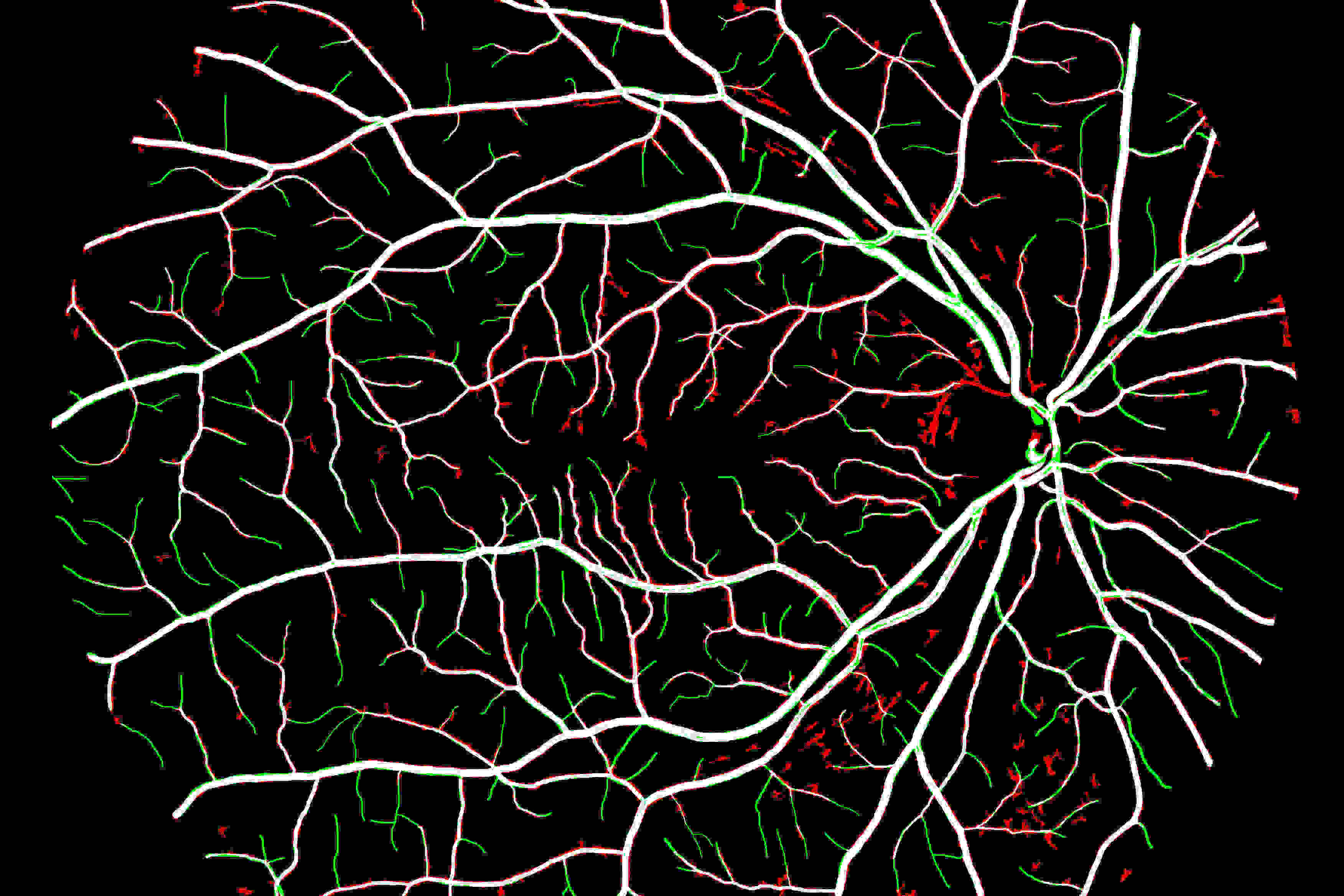}
	\caption{ours}\label{fig:seg:proposed}
	\end{subfigure}
	\vspace{-10pt}
	\caption{Vessel segmentation results obtained by the proposed and the state-of-the-art vessel-like structures enhancement methods followed by the same local thresholding approach proposed in~\cite{jiang2003adaptive} when applied to the HRF dataset images.
		(\subref{fig:seg:vessel})~vesselness,
		(\subref{fig:seg:clahe})~CLAHE,
		(\subref{fig:seg:soth})~Zana's top-hat,
		(\subref{fig:seg:neurite})~neuriteness,
		(\subref{fig:seg:pctv})~PCT vesselness,
		(\subref{fig:seg:pctn})~PCT neuriteness,
		(\subref{fig:seg:wavelet})~wavelet,
		(\subref{fig:seg:bcosfire})~line detector,
		(\subref{fig:seg:vr})~volume ratio, and (\subref{fig:seg:proposed})~the bowler-hat. Colours indicate true positive (white), false positive (red) and false negative pixels (green). Corresponding mean AUC values can be found in~\Cref{tab:acc}.}\label{fig:seg}
%\vspace{-10pt}
\end{figure*}
%TABLE
\begin{table}[t!]
	\centering
	\resizebox{0.5\linewidth}{!}{%
	\begin{tabular}{llc}
		\toprule
		\multirow{2}{2.5cm}{\centering Enhancement Method}&\multicolumn{2}{c}{\centering ACC (std)}\\ \cmidrule(l){2-3}
						 &Year/Ref\quad &HRF  \quad\\
		\midrule
		Vesselness		&1998~\cite{frangi1998multiscale}	&0.951(0.006) \\
		\midrule
		CLAHE			&1998~\cite{Pisano1998}				&0.859(0.009) \\
		\midrule
		Zana's top-hat 		&2001~\cite{ZK2001}					&0.946(0.008) \\
		\midrule
		Neuriteness		&2004~\cite{meijering2004design}	&0.953(0.006) \\
		\midrule
		PCT vesselness 	&2012~\cite{obara2012contrast}		&0.926(0.007) \\
		\midrule
		PCT neuriteness	&2012~\cite{obara2012contrast} 		&0.900(0.008) \\
		\midrule
		Wavelet			&2012~\cite{bankhead2012fast}	 	&0.946(0.006) \\
		\midrule
		Line detector	&2013~\cite{nguyen2013effective}	&0.957(0.006) \\
		\midrule
		Volume ratio	&2016~\cite{JPLetal2016} 			&0.947(0.011) \\
		\midrule
		Bowler-hat		&-									&\textbf{0.961(0.005)} \\
		\thickhline\thickhline
	\end{tabular}
}
	\caption{Mean ACC values with the standard deviation for vessel segmentation results obtained by the proposed and the state-of-the-art vessel-like structures enhancement methods followed by the same local thresholding approach proposed in~\cite{jiang2003adaptive} when applied to the HRF dataset images.}\label{tab:acc}
\end{table}

%TABLE
\begin{table}[t!]
	\centering
	\resizebox{1\linewidth}{!}{%
	\begin{tabular}{lccc|ccc|ccc}
		\thickhline
		\multirow{2}{*}{Method} &\multicolumn{3}{c}{DRIVE}&\multicolumn{3}{c}{STARE}&\multicolumn{3}{c}{HRF}\\ \cmidrule(lr){2-4}  \cmidrule(lr){5-7} \cmidrule(lr){8-10}
						 						&SE &SP &ACC  &SE &SP &ACC &SE &SP &ACC \\ \hline
		Staal et.al~\cite{staal2004}			&- &- &0.946 &- &- &0.951 &-&-&-\\ \hline
		Soares et.al~\cite{soares2006retinal}	&- &- &0.946 &- &- &0.948 &-&-&- \\ \hline
		Lupascu et.al~\cite{lupascu2010fabc}	&0.720 &- &\textbf{0.959} &- &- &- &-&-&- \\ \hline
		You et.al~\cite{you2011segmentation}	&0.741 &0.975 & 0.943 &0.726 &0.975 &0.949 &-&-&- \\ \hline
		Marin et.al~\cite{marin2011new}			&0.706 &0.980 &0.945 &0.694 &0.981 &0.952 &-&-&- \\ \hline
		Wang et.al~\cite{wang2013retinal} 		&- &- &0.946 &- &- &0.952 &-&-&- \\ \hline
		Mendonca et.al~\cite{mendonca2006segmentation}		&0.734 & 0.976 &0.945 &0.699 &0.973 &0.944 &-&-&- \\ \hline
		Palomera-Perez et.al~\cite{palomera2010parallel}	&0.660 &0.961 &0.922 &\textbf{0.779} & 0.940 &0.924  &-&-&-\\ \hline
		Matinez-Perez et.al~\cite{martinez2007segmentation} &0.724 &0.965 &0.934 & 0.750 &0.956 &0.941 &-&-&- \\ \hline
		Al-Diri et.al~\cite{al2009active}		&0.728 & 0.955 &- & 0.752 &0.968 &- &-&-&- \\\hline
		Fraz et.al~\cite{FBRetal2012}			&0.715 &0.976 &0.943 &0.731 &0.968 &0.944 &-&-&-\\ \hline
		Nguyen et.al~\cite{nguyen2013effective}		&- &- &0.940 &- &- &0.932 &-&-&- \\ \hline
		Bankhead et.al~\cite{bankhead2012fast}		&0.703 &0.971 &0.937 &0.758 & 0.950 &0.932 &-&-&-\\ \hline	
		Orlando et.al~\cite{orlando2014learning}		&\textbf{0.785} &0.967 &- &- &- &- &-&-&-\\ \hline
		Azzopardi et.al~\cite{azzopardi2015trainable}	&0.766 &0.970 &0.944 &0.772 &0.970 &0.950 &-&-&-\\ \hline
		Odstrcilik et.al~\cite{odstrcilik2013retinal}  &0.784 &0.951 &0.934 &0.706 &0.969 &0.934 &0.786  &0.975 &0.953 \\ \hline
		Zhang et.al~\cite{zhang2016robust}  &0.774 &0.972 &0.947 &0.779 &0.975 &0.955 &0.797  &0.971 &0.955 \\ \hline
		Proposed method								  &0.718 &\textbf{0.981} &\textbf{0.959} &0.730 &\textbf{0.979} &\textbf{0.962} &\textbf{0.831} &\textbf{0.981} &\textbf{0.963}\\ \hline
		\thickhline \thickhline
	\end{tabular}}
	\caption{Performance of different vessel segmentation methods have been reported in the literature with the proposed method, regarding mean sensitivity (SE), specificity (SP), accuracy (ACC) on the all over the DRIVE, STARE and  HRF datasets.}\label{tab:seg}
\end{table}

\subsubsection{Comparison with Other Segmentation Methods}\label{subsubsec:seg-comparion}

To highlight the effectiveness of the proposed vessel enhancement method (combined with the local thresholding approach~\cite{jiang2003adaptive}) for a full vessel segmentation, we compared the performance of our method with seventeen state-of-the-art vessel segmentation methods reported in the literature~\cite{staal2004,soares2006retinal,lupascu2010fabc, you2011segmentation,marin2011new,wang2013retinal,mendonca2006segmentation,palomera2010parallel,martinez2007segmentation,al2009active,FBRetal2012,nguyen2013effective,bankhead2012fast,orlando2014learning,azzopardi2015trainable,zhang2016robust,odstrcilik2013retinal} applied to DRIVE, STARE and HRF datasets.

	~\Cref{tab:seg} shows the reported results of the seventeen segmentation methods compared with the proposed method. From~\Cref{tab:seg}, it can be seen that the proposed bowler-hat transform outperforms several common or state-of-the-art methods from the field. In cases where the proposed method does not outperform, but still performs to a similar quality, it is worth keeping in mind that many of these methods combine multiple stages, of which enhancement is just one, whereas our approach is able to achieve such high quality results with just an enhancement process.
The results on both datasets demonstrate that the sensitivity of the proposed method is not in top three respectively for DRIVE ($SE=0.616$) and STARE ($SE=0.730$). However, the proposed method has the highest score with the specificity ($SP=0.991$) for DRIVE and ($SP=0.979$) for STARE. Most importantly, our method has the accuracy ($ACC=0.960$) and ($ACC=0.962$) for DRIVE and STARE respectively; the highest compared to other vessel segmentation methods. Finally, the proposed method has the highest score for HRF dataset, with ($SE=0.831$), ($SP=0.981$) and ($ACC=0.963$).

\subsection{Other Biomedical Data}\label{subsubsec:other}
While we have demonstrated the proposed method on the enhancement of vessel-like structures, the approach is feasible for a wide range of biomedical images, see~\Cref{fig:other}.
% FIGURE
\renewcommand\sfigure{0.3}
\begin{figure}[h!]
	\centering
	\begin{subfigure}{\sfigure\linewidth}
		\includegraphics[width=0.45\linewidth, clip, trim=100mm 105mm 35mm 120mm]{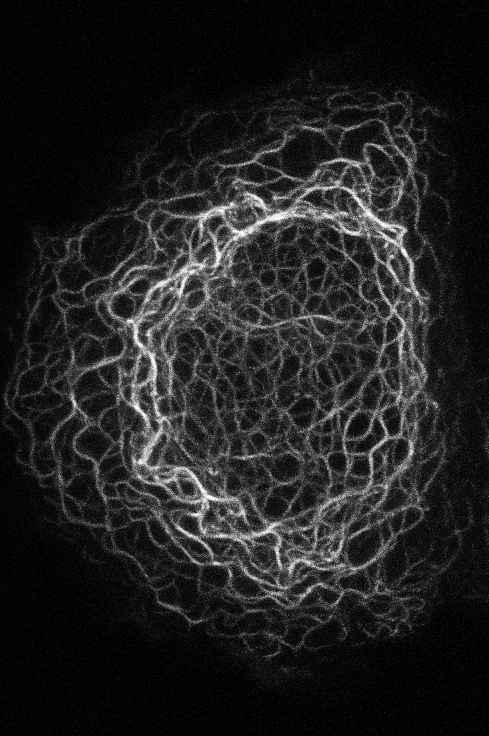}
		\includegraphics[width=0.45\linewidth, clip, trim=100mm 105mm 35mm 120mm]{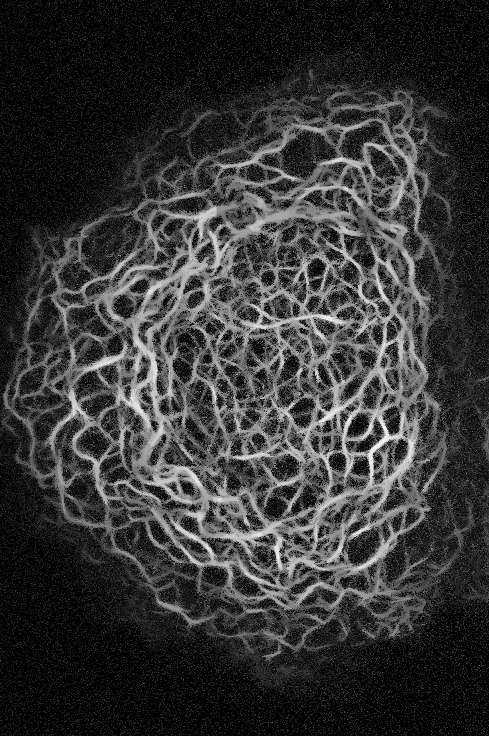}
		\caption{\quad}\label{fig:other:skin}
	\end{subfigure}
	\begin{subfigure}{\sfigure\linewidth}
		\includegraphics[width=0.45\linewidth, clip, trim=40mm  25mm 65mm 60mm]{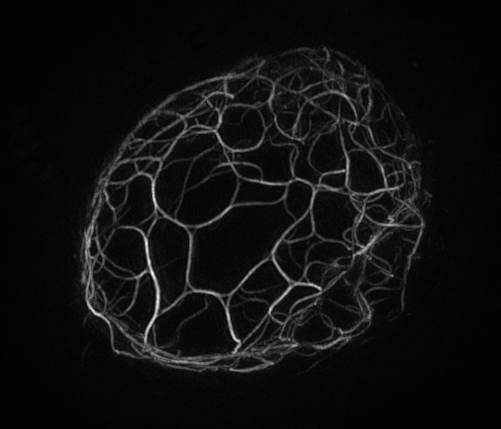}
		\includegraphics[width=0.45\linewidth, clip, trim=40mm 25mm 65mm 60mm]{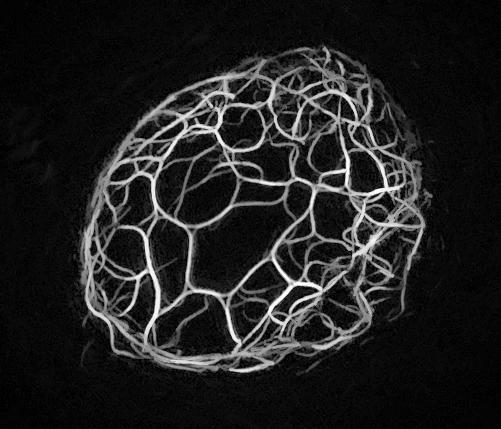}
		\caption{\quad}\label{fig:other:keratin}
	\end{subfigure}
	\begin{subfigure}{\sfigure\linewidth}
		\includegraphics[width=0.45\linewidth, clip, trim=60mm 65mm 60mm 60mm]{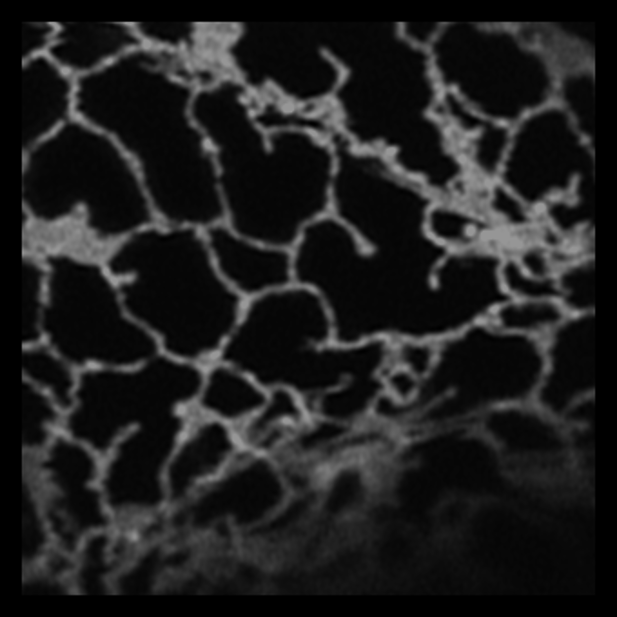}
		\includegraphics[width=0.45\linewidth, clip, trim=60mm 65mm 60mm 60mm]{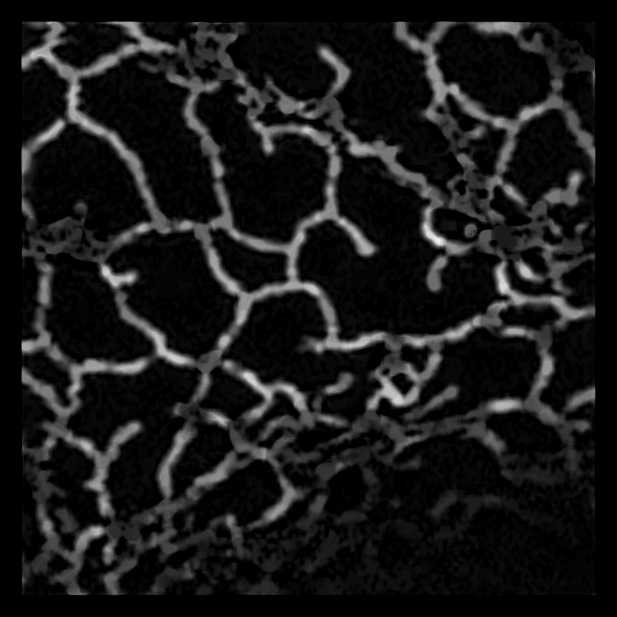}
		\caption{\quad}\label{fig:other:er}
	\end{subfigure}
	\\ \vspace{5pt}
	\begin{subfigure}{\sfigure\linewidth}
		\includegraphics[width=0.45\linewidth, clip, trim=0mm 25mm 50mm 60mm]{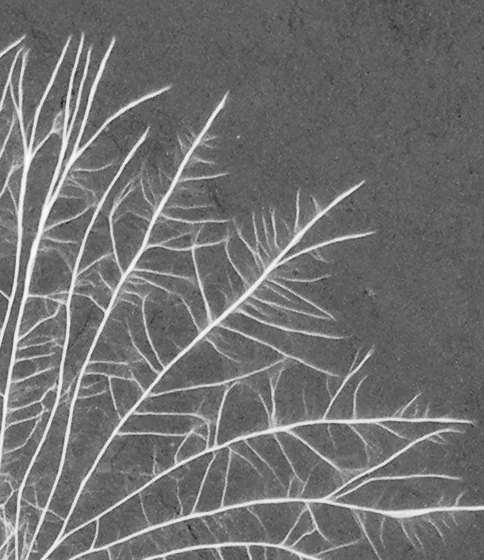}
		\includegraphics[width=0.45\linewidth, clip, trim=0mm 25mm 50mm 60mm]{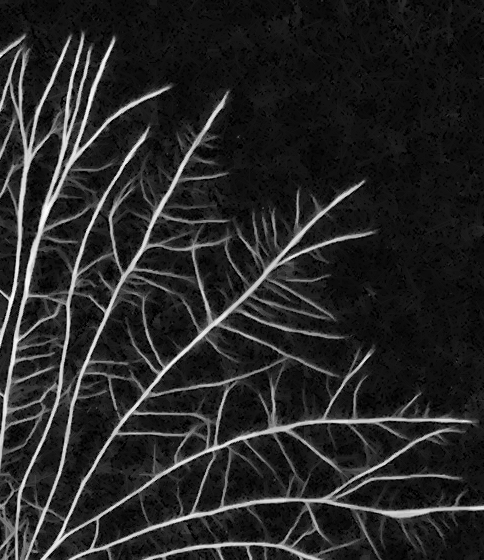}
		\caption{\quad}\label{fig:other:fungal}
	\end{subfigure}
	\begin{subfigure}{\sfigure\linewidth}
		\includegraphics[width=0.45\linewidth, clip, trim=35mm 20mm 0mm 5mm]{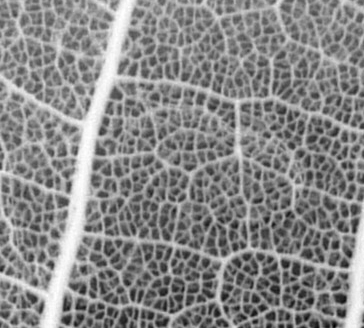}
		\includegraphics[width=0.45\linewidth, clip, trim=35mm 20mm 0mm 5mm]{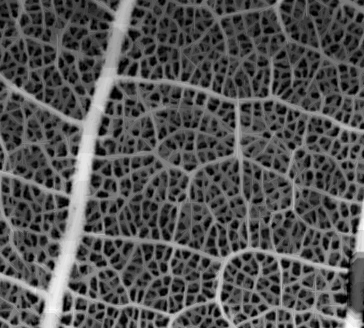}
		\caption{\quad}\label{fig:other:leaf}
	\end{subfigure}
	\caption{Results of the vessel-like structure enhancement using the bowler-hat on biological images of (\subref{fig:other:skin}--\subref{fig:other:keratin}) cytoskeletal networks, (\subref{fig:other:er}) endoplasmic reticulum, and (\subref{fig:other:fungal}--\subref{fig:other:leaf}) macro-scale networks . (\subref{fig:other:skin})~provided by Prof. R. Leube, RWTH Aachen University, Germany. (\subref{fig:other:keratin})~ provided by Dr T. Hawkins, Durham University, UK. (\subref{fig:other:er}--\subref{fig:other:leaf})~provided by Prof. M. Fricker, Oxford University, UK.}\label{fig:other}
\end{figure}

%------------------------------------------------------------------------------%
% !TEX root = ../main-elsevier.tex
\section{Conclusion and Discussion}\label{sec:Conclusion}

A wide range of image processing methods have been proposed for vessel-like structure enhancement in biomedical images, see section~\Cref{sec:intro}. Most of them, however, suffer from issues with low-contrast signals, enhancement of noise or when dealing with junctions.

In this paper, we introduce a new enhancement method for vessel-like structures based on mathematical morphology, which exploits the elongated shape of vessel-like structures. The proposed method, the bowler-hat transform, was qualitatively and quantitatively validated and compared with the state-of-the-art methods using a range of synthetic and real image datasets, including retinal image collections (DRIVE, STARE and HRF). We showed the effectiveness of the bowler-hat transform, and its superior performance on retinal imaging data, see~\Cref{fig:roc},~\Cref{tab:auc}, and~\Cref{tab:acc}. Furthermore, experimental results on the unhealthy retinal images have shown that the vessels enhanced by our bowler hat transform are continuous and complete in problematic regions as illustrated in~\Cref{fig:unhealthy}.

As with any image processing technique, our proposed method has limitations. Basically, morphological operations are renowned for their large computational requirements. 
Another limitation of the proposed method is displayed in~\Cref{fig:compare} row 4, which shows a vessel-like structure with an attached `blob' (green arrow), a perfect vessel enhancement method would enhance all of the linear structure and none of the blob. Whilst none of the comparison methods act in this ideal manner many of them show a clear difference between the blob response and vessel response, our proposed method shows some difference, but this difference impacts the signal of the vessel.

Moreover, as we note in~\Cref{fig:hrf}, the proposed method is sensitive to noise such as susceptible to speckle and salt\&pepper, as is the PCT neuriteness method in~\Cref{fig:hrf:pctneurite}. In the future, we will investigate introducing a line-shaped morphological structuring element with varying thickness to address this issue. Nevertheless, our implementation demonstrates an improved and easy to use vessel enhancement alternative that can be used in a wide range of biomedical imaging scenarios~\cite{trzupek2011intelligent}.
Whilst one would expect the lack of noise suppression to be a major issue with regard to quantified measurements of vessel enhancement, we find that the proposed method gives the best enhancement of all methods on the DRIVE, STARE and HRF datasets (see~\Cref{tab:auc} and~\Cref{fig:roc}).

Future extensions of this work will include the development of a three-dimensional equivalent, exploration of blob-like structures enhancing variants of this method, and an analysis of parameter sensitivity for different modalities.
 
\section{Acknowledgement}
\c{C}i\u{g}dem~Sazak and Carl~J.~Nelson contributed equally to this work. \c{C}i\u{g}dem Sazak is funded by the Turkey Ministry of National Education. Carl~J.~Nelson is funded by EPSRC UK (1314229); and the project was supported by a Royal Society UK (RF080232).
% \vskip-10px
\section*{References}
%------------------------------------------------------------------------------%
\bibliography{bib/Reference}
\newpage
%------------------------------------------------------------------------------%

\noindent {\bf Cigdem Sazak} received her B.S. degree in Computer Engineering from Sakarya University, Turkey, in 2011. She received his MSc degree in 2014 from Leicester University and is currently a PhD student in the Department of Computer Science at the University of Durham. Her research interests are in the image enhancement of biomedical images.

\noindent {\bf Carl J Nelson} received a PhD in Computer Science from Durham University, UK in 2017. His research focused on the use of mathematical morphology for the analysis of bioimages. He is currently a research associate at the University of Glasgow, UK, where his interests focus on the interface of software and hardware for improved acquisition in light sheet microscopy.

\noindent{\bf Boguslaw Obara}  received a PhD in Computer Science from AGH, Poland. He has been a Fulbright fellow and a postdoctoral researcher at University California, USA and University of Oxford, UK. He is currently an associate professor in Computer Science at Durham University, UK. His research focuses on image processing techniques.
%----------------------------------------------------------------------%

\end{document}